\documentclass[10pt,journal,compsoc]{IEEEtran}

\usepackage{epsfig}
\usepackage{graphicx}
\usepackage{amsmath}
\usepackage{amssymb}

\usepackage{algpseudocode}
\usepackage{cases}
\usepackage{subfigure}
\usepackage{overpic}
\usepackage{multirow}
\usepackage{color}
\usepackage{array}
\usepackage{tabularx}
\usepackage[linesnumbered,ruled]{algorithm2e}
\usepackage{bbding}
\definecolor{mypink}{RGB}{219, 48, 122}

\usepackage{booktabs}
\usepackage{hhline}
\usepackage{makecell}
\usepackage{colortbl}
\usepackage[normalem]{ulem}
\usepackage{soul}
\usepackage{url}
\definecolor{Gray}{rgb}{0.7,0.7,0.7}

\newcolumntype{x}{>\small c}

\newcolumntype{o}{>\small L}

\ifCLASSOPTIONcompsoc
  \usepackage[nocompress]{cite}
\else
  \usepackage{cite}
\fi
\ifCLASSINFOpdf
\else
\fi
\begin{document}
%
\title{Learning to Adapt Invariance in Memory for Person Re-identification}
%
%
%
%

\author{Zhun Zhong,
        Liang Zheng,
        Zhiming Luo,
        Shaozi Li,
        Yi Yang
\IEEEcompsocitemizethanks{
\IEEEcompsocthanksitem Z. Zhong is with the Department of Artificial Intelligence, Xiamen University,
Xiamen 361005, China and also with the Centre for Artificial Intelligence, University of Technology Sydney, Ultimo, NSW 2007, Australia (e-mail: zhunzhong007@gmail.com).
\IEEEcompsocthanksitem L. Zheng is with the Research School of Computer Science, The Australian National University, Canberra, ACT 0200, Australia (e-mail: liangzheng06@gmail.com).
\IEEEcompsocthanksitem Z. Luo is with the Postdoc Center of Information and Communication Engineering, Xiamen University, Xiamen 361005, China (e-mail: zhiming.luo@xmu.edu.cn).
\IEEEcompsocthanksitem S. Li (corresponding author) is with the Department of Artificial Intelligence, Xiamen University, Xiamen 361005, China (e-mail: szlig@xmu.edu.cn).
\IEEEcompsocthanksitem Yi Yang is with the Centre for Artificial Intelligence, University of Technology Sydney, Ultimo, NSW 2007, Australia. E-mail: yee.i.yang@gmail.com
}}

%
%

\markboth{Journal of \LaTeX\ Class Files, 2019}%
{Shell \MakeLowercase{\textit{et al.}}: Bare Demo of IEEEtran.cls for Computer Society Journals}
%



\IEEEtitleabstractindextext{%
\begin{abstract}
This work considers the problem of unsupervised domain adaptation in person re-identification (re-ID), which aims to transfer knowledge from the source domain to the target domain.
Existing methods are primary to reduce the inter-domain shift between the domains, which however usually overlook the relations among target samples.
This paper investigates into the intra-domain variations of the target domain and proposes a novel adaptation framework \textit{w.r.t} three types of underlying invariance, \textit{i.e.}, Exemplar-Invariance, Camera-Invariance, and Neighborhood-Invariance.
Specifically, an exemplar memory is introduced to store features of samples, which can effectively and efficiently enforce the invariance constraints over the global dataset.
We further present the Graph-based Positive Prediction (GPP) method to explore reliable neighbors for the target domain, which is built upon the memory and is trained on the source samples.
Experiments demonstrate that 1) the three invariance properties are indispensable for effective domain adaptation, 2) the memory plays a key role in implementing invariance learning and improves the performance with limited extra computation cost, 3) GPP could facilitate the invariance learning and thus significantly improves the results, and 4) our approach produces new state-of-the-art adaptation accuracy on three re-ID large-scale benchmarks.

\end{abstract}

}

\maketitle

\IEEEdisplaynontitleabstractindextext

%
\IEEEpeerreviewmaketitle

\IEEEraisesectionheading{\section{Introduction}\label{sec:introduction}}

%
%
%
%
\IEEEPARstart{P}{erson} re-identification (re-ID) \cite{zheng2016personsurvery} is an image retrieval task, that aims at seeking matched persons of the query from a disjoint-camera database. 
The predominant methods have demonstrated dramatic performance when trained and tested on the same data distribution. However, they may suffer a significant degradation in the performance when evaluated on a different domain, due to dataset shifts from the changes of scenario, season, illumination, camera deployment, et al. 
It raises a domain adaptation problem that often encountered in real world applications and attracts increasing attention in the community \cite{fan2017pul,deng2018image,wang2018reid,Zhong_2018_ECCV,zhong2019invariance}. 
In this work, we study the problem of unsupervised domain adaptation (UDA) in re-ID. The goal is to improve the generalization ability of models on a target domain, using a labeled source domain and an unlabeled target domain.

Conventional methods of UDA are mainly designed for a closed-set setting, where the source and target domains share a common label space, \textit{i.e.} the classes of two domains are exactly the same. A popular approach is to align the feature distributions of both domains, but it does not readily apply to the context of re-ID. Since domain adaptation in re-ID is a special open-set problem \cite{busto2017open-set,saito2018open,sohn2019unsupervised}, where the source and target domains have completely disjoint classes/identities. For such label constraint, directly aligning the feature distributions of two domains will align the samples from different classes and may be detrimental to the adaptation accuracy.

To address the challenges of domain adaptive re-ID, recent works concentrate on aligning the source-target distributions in a common space, such as pixel-level space \cite{deng2018image,wei2018person} and attribute label space \cite{wang2018reid,lin2018multibmvc}. Despite their success, these works only consider the overall inter-domain shift between the source and target domains, but largely overlook the intra-domain variations of the target domain. In the re-ID system, the intra-domain variations are important factors that affect the performance. 
Without considering the intra-domain variations of the target domain, an adapted model will produce poor performance, when the intra-domain variations in the target testing set are seriously different from the source domain.

In this work, we explicitly consider the intra-domain variations of the target domain and design our framework \textit{w.r.t} three types of underlying invariance, \textit{i.e.}, Exemplar-Invariance (EI), Camera-Invariance (CI), and Neighborhood-Invariance (NI), as described below.

\textbf{Exemplar-Invariance (EI)}: The first property is motivated by the retrieval results of re-ID. Given a re-ID model trained on a labeled source training set, we evaluate it on a source/target testing set. On the one hand, we observe that the top-ranked retrieval results (both positive and negative samples) always are visually similar to the query when tested on the source set. A similar phenomenon is shown in image classification \cite{wu2018unsupervised}. This indicates that the model has learned to distinguish persons by apparent similarity for the source domain. On the other hand, when tested on the target set, the top-ranked results often include many samples that are visually dissimilar to the query. This suggests that the ability of the model to distinguish persons by apparent similarity is degraded on the target domain. In reality, each person exemplar could differ significantly from others even shared the same identity. Therefore, it is possible to enable the model to capture the apparent representation by learning to distinguish individual exemplars. 
To achieve this goal, we introduce the exemplar-invariance (EI) to improve the discrimination ability of the model on the target domain, by encouraging each exemplar to be close to itself while away from others.

\textbf{Camera-Invariance (CI)}: Camera style (CamStyle) difference is a critical factor for re-ID that can be clearly identified, since the appearance of a person may change largely under different cameras \cite{zhong2018camera,zhong2019camstyle}. Due to the camera deployments of the source and target domains are usually different, the model trained on the source domain may suffer from the variations caused by the target cameras. %
To address this problem, Zhong \emph{et al.} \cite{Zhong_2018_ECCV} introduce camera-invariance (CI) by enforcing an target example and its corresponding CamStyle transferred images to be close to each other. Inspired by them, we integrate the camera-invariance learning into our model by classifying an target example and its CamStyle counterparts to a same class.

\begin{figure}[!t]
    \centering
    \includegraphics[width=0.98\linewidth]{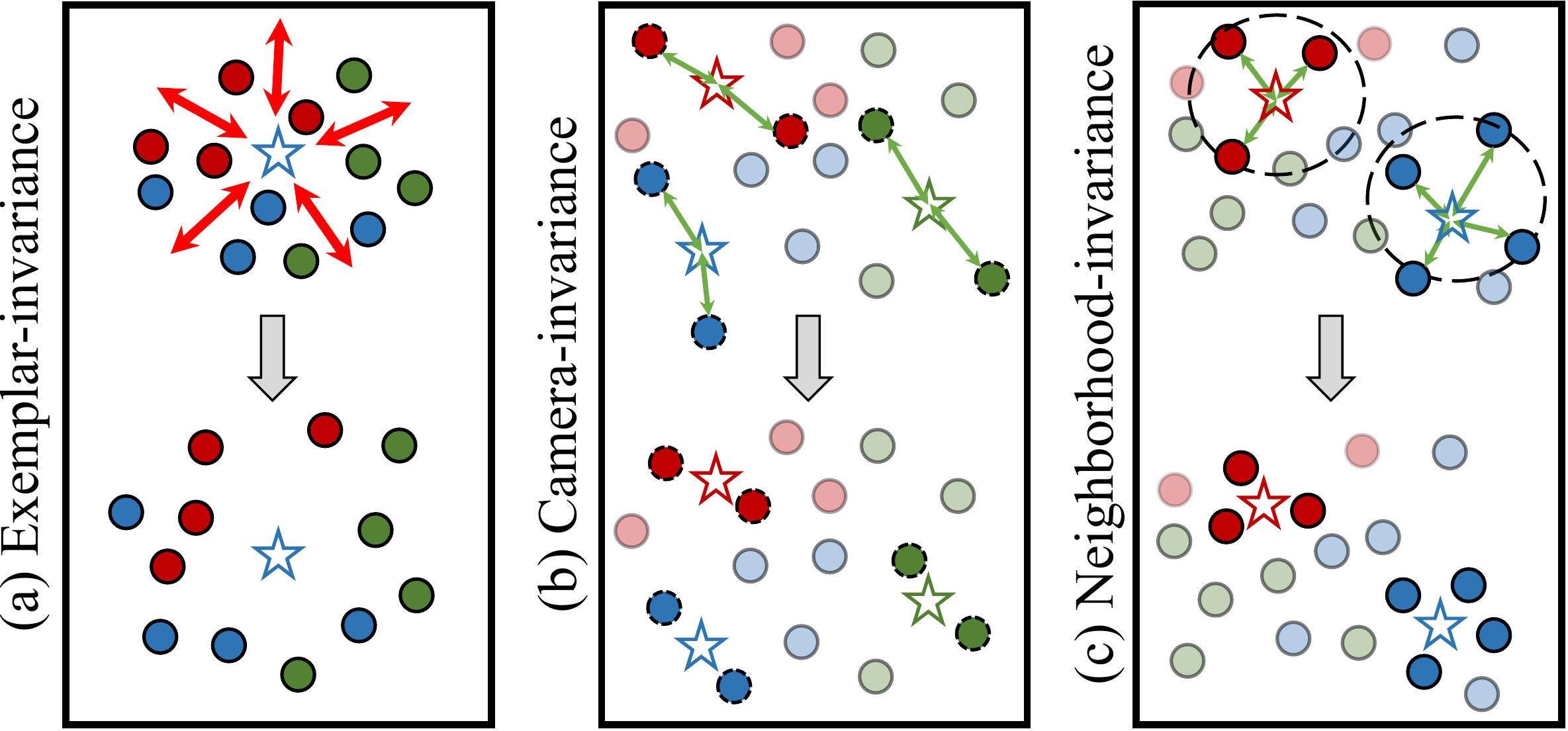}
    \caption{Examples of three underlying properties of invariance. Colors indicate identities. (a) Exemplar-invariance: an input exemplar (denoted by $\star$) is enforced to be away from others. (b) Camera-invariance: an input exemplar (denoted by $\star$) and its CamStyle transferred images (with dashed outline) are encouraged to be close to each other. (c) Neighborhood-invariance: an input exemplar (denoted by $\star$) and its reliable neighbors (highlighted in dashed circle) are forced to be close to each other. Best viewed in color.}
    \label{fig:three_invariance}
\end{figure}

\textbf{Neighborhood-Invariance (NI)}: Apart from the easily identified camera variance, some other latent intra-domain variations are hard to explicitly discern without fine-grained labels, such as the changes of pose, view, and background. To overcome this difficulty, we attempt to generalize the model with the neighbors of target samples. Suppose we are given an appropriate model trained on the source and target domains, a target sample and its nearest-neighbors in the target set may share the same identity with a higher potential. Considering this trait, we introduce the  neighborhood-invariance (NI) to learn a model that is more robust to overcome the latent intra-domain variations of the target domain. We accomplish this constraint by encouraging an exemplar and its reliable neighbors to be close to each other. Examples of the three types of invariance are illustrated in Fig.~\ref{fig:three_invariance}.

Intuitively, a straightforward way to implement the three invariance properties is to constrain them with contrastive/triplet loss \cite{hadsell2006contrastive,hermans2017defense} within a training mini-batch. However, the number of samples in a mini-batch is relatively small compared with the entire training set. In this manner, it is difficult to form a mini-batch with ideal examples, and the overall relations between training samples cannot be considered thoroughly during the network adaptation procedure. To deal with this issue, we develop a novel framework to effectively accommodate the three invariance properties for domain adaptive re-ID. Specifically, we introduce an exemplar memory module into the network to store the up-to-date representations of all training samples. The memory enables the network to enforce the invariance constraints over the entire/global target training data instead of the current mini-batch. With the memory, the invariance learning of the target domain can be effectively implemented with a non-parametric classification loss, considering each target sample as an individual class. 

In our previous work \cite{zhong2019invariance}, we directly select top-$k$ nearest neighbors from the memory for the learning of NI. This straightforward strategy ignores the underlying relations between samples in the memory. As a result, the similarity estimation of hard samples may not be accurate when the model has inferior discriminative ability. As a notable extension of our previous work \cite{zhong2019invariance}, we propose a graph-based positive prediction (GPP) approach to address this problem, thereby promoting the invariance learning. GPP is built upon the memory module and designed by graph convolutional networks (GCNs), which aims to predict positive neighbors from the memory for a training target sample. In addition to the target memory, we also construct a memory for saving features of the source domain. This enables us to imitate the neighbor exploring process of target invariance learning and thus learn GPP on the labeled source domain. The learned GPP is then directly applied to the unlabeled target domain for facilitating the learning of NI.

In summary, our contribution is as follow:

\begin{itemize}
\item This work comprehensively investigates the intra-domain variations of the target domain and studies three underlying properties of target invariance. The experiment demonstrates that the three properties are indispensable for improving the transferable ability of the model in the context of re-ID.

\item This work proposes a novel framework equipped with a memory module that can effectively enforce the three constraints into the network. The memory enables us to fully exploit the sample relations over the whole training set instead of the mini-batch. With the memory, the performance can be significantly improved, requiring very limited extra computation cost and GPU memory.

\item This work introduces a Graph-based Positive Prediction (GPP) approach to leverage the relationships between candidate neighbors and infer accurate positive neighbors for the training target sample. The experiment shows that GPP is beneficial to the learning of neighborhood-invariance and could consistently improve the results, especially the mAP. 

\item In addition, we analyze the mechanism of the three invariance properties, which helps us to understand how these three properties encourage the network to adapt to the target domain.

\item Experiments demonstrate the effectiveness and superiority of the proposed method over the state-of-the-art UDA approaches. Our results outperform state of the art by a large margin on three datasets, including Market-1501, DukeMTMC-reID and MSMT17.

\end{itemize}

\section{Related Work}

\textbf{Unsupervised domain adaptation.} This work is closely related to unsupervised domain adaptation (UDA). In classical UDA, methods are designed under the assumption of the closed-set scenario, where the classes of the source and target domains are precisely the same. For this problem, recent popular approaches are mainly focused on distribution alignment learning, which attempts to reduce the domain discrepancy by using Maximum Mean Discrepancy (MMD) minimization \cite{gretton2007kernel,long2015learning,yan2017mind} or domain adversarial training \cite{bousmalis2016domain,tzeng2017adversarial}. However, these methods are usually not applicable to the open-set scenario, where unknown classes exist in the source/target domain. Since samples of unknown classes should not be aligned with the ones of known classes. It raises to the problem of open set UDA, introduced by Busto and Grall \cite{busto2017open-set}. To tackle this problem, Busto and Grall \cite{busto2017open-set} develop a method to learn a mapping from the source domain to the target domain by jointly predicting unknown-class target samples and discarding them. Saito \textit{et al.} \cite{saito2018open} introduce an adversarial learning framework to separate target samples into known and unknown classes. Meanwhile, unknown classes are rejected during feature alignment. Recently, Sohn \textit{et al.} \cite{sohn2019unsupervised} consider a more challenging setting of open-set UDA, where the source and target domain belong disjoint label spaces, \textit{i.e.} the classes of two domains are completely different. In practice, many tasks match such setting, \textit{e.g.}, cross-ethnicity face recognition and UDA in re-ID considered in our work. To address this problem, Sohn \textit{et al.} first reformulate the disjoint classification task to a binary verification one. Then, a Feature Transfer Network (FTN) is proposed to transfer the source feature space to a new space and align with the target feature space. In this paper, we study this problem in the aspect of intra-domain variations in the target domain and propose an effective framework to improve the generalization ability of the model by target invariance learning.

\textbf{Unsupervised person re-identification (re-ID).} 
Recent supervised methods have made great achievements in re-ID \cite{Li_2018_CVPR,zhong2017re,sun2018beyond,zheng2019joint,sun2019dissecting}, benefiting from rich-labeled data and the increasing ability of deep networks \cite{Yawei2019Taking,resnet}. However, the performance of these strong methods may have a large drop when tested on an unseen (target) dataset, due to the dataset shift. To address this problem, many hand-craft features based methods are designed for unsupervised re-ID, such as ELF \cite{gray2008viewpoint}, LOMO \cite{liao2015lomo} and BOW \cite{zheng2015scalable}. These methods can be directly applied to any dataset without training, but fail to obtain fulfilling performance on large-scale, complex scenarios. Although labeling re-ID data is difficult, it is relatively easy to collect sufficient unlabeled data in the target domain. Recently, many works are proposed to transfer the knowledge of a labeled data to an unlabeled one. These works mainly can be divided into two categories: 1) discovering pseudo labels for target dataset, and 2) reducing the source-target discrepancy in a common label space. 
For the first category, methods use a labeled source dataset to initialize a re-ID model and explore pseudo labels for target dataset based on clustering \cite{yu2017cross, fan2017pul}, associating label with labeled source dataset \cite{yu2019unsupervised}, assigning label with nearest-neighbors\cite{chen2018deep, yang2018leveraging,li2018unsupervised,li2019unsupervised}, or regarding camera style counterparts as positive sample \cite{Zhong_2018_ECCV}. These methods are closely related to our work in that using the relationship between target samples to refine the model. The main difference is that our work comprehensively considers three latent invariance constraints and enforces them over the entire dataset. We show the mutual benefit among the three invariance properties and the effectiveness of exploiting the global sample relationship.
Methods of the second category attempt to align the distributions of the source and target domains in a common space, such as image-level space \cite{deng2018image,wei2018person,Bak_2018_ECCV,zhong2019camstyle} and attribute label space \cite{wang2018reid,lin2018multibmvc}. These methods only consider the overall discrepancy between the source and target domains, but largely ignore the intra-domain variations in the target domain. In this work, we explicitly consider the invariance properties in the target domain to address the problem of domain adaptive re-ID.

\textbf{Neural networks with augmented-memory.} Learning neural networks with augmented-memory is proposed to address various tasks, such as question answering \cite{MemoryNetworks2015,sukhbaatar2015end}, few-shot learning \cite{santoro2016meta,vinyals2016matching}, and video understanding \cite{wu2018long}. The augmented-memory enables the networks to store the intermediate knowledge into a structural and addressable table. The community of learning with augmented-memory mainly can be divided into two categories. The first category aims to augment neural networks with a fully differentiable memory module, such as Neural Turing Machine \cite{graves2014neural} and Memory Networks \cite{MemoryNetworks2015}. Memory Networks introduce a long-term memory component that can be read and written. The memory is utilized to query and retrieve fact for the task of question answering. Another category focus on developing a non-parametric memory \cite{vinyals2016matching,xiao2017joint,wu2018unsupervised,wu2018improving} which can directly save the features of samples into a feature bank and be updated during training. These approaches then use the attention mechanism to calculate the similarities between the query and instances in the memory. We draw inspiration from these methods and develop a memory-based framework for unsupervised domain adaptive re-ID.

\begin{figure*}[!t]
    \centering
    \includegraphics[width=0.98\linewidth]{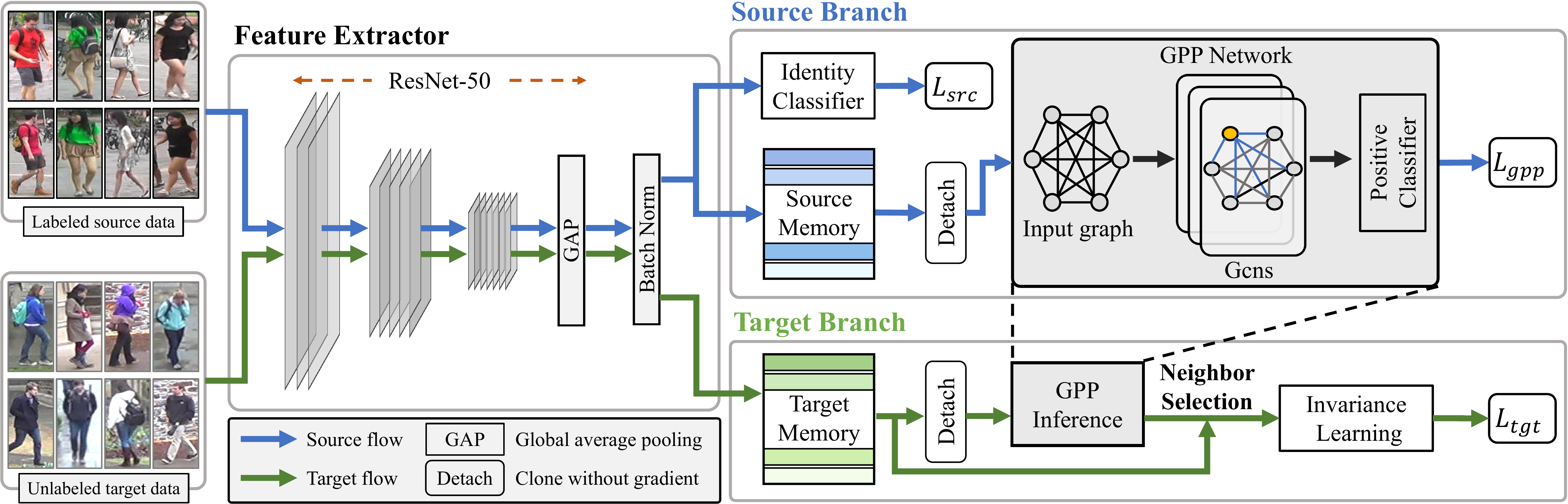}
    \caption{The framework of the proposed method. During training, the inputs are drawn from the labeled source domain and the unlabeled domain. We feed-forward the inputs into the feature extractor to obtain the up-to-date representations. Subsequently, two branches are designed to optimize the framework with the source data and the target data, respectively. In each branch, we introduce a memory to store the features of corresponding domain data. The source branch aims to learn basic representation for the model with identity classification loss $L_{scr}$, as well as, to learn a graph-based positive prediction (GPP) network with binary classification loss $L_{gpp}$. For training of GPP, we first select top-ranked features of the input from the source memory. Then, the selected features are regarded as candidate neighbors and are used to train the GPP network. The GPP network is trained to predict the positive and negative neighbors of the input. The target branch attempts to enforce the invariance learning on the target data. The invariance learning loss $L_{tgt}$ is calculated by estimating the similarities between the target sample and whole features in the target memory. In addition, we employ the GPP network to infer reliable positive neighbors for the target sample, thereby facilitating the invariance learning. In testing, the L2-normalized output of the global average pooling (GAP) is used as the feature of an image.}
    \label{fig:framework}
\end{figure*}

\textbf{Graph convolutional network (GCN)}.
In real-world applications, data are usually represented in the form of graphs or networks, such as knowledge graphs and social networks. Recently, many works concentrate on extending the deep convolutional neural network for handling graph data. In light of the objectives on graph data, these works are generally divided into spectral based methods \cite{bruna2014spectral,defferrard2016convolutional,henaff2015deep,kipf2016semi} and spatial based methods \cite{monti2017geometric,niepert2016learning,hamilton2017inductive,shen2018graph,wang2019linkage}. The spectral based methods are developed based on graph spectral theory. These methods usually handle the entire graph to learn a domain-dependent graph network, leading them difficult to apply on large graphs and can not be applied to a different structure.
The spatial based methods extend the convolution operation of CNNs to the graph structure. The graph convolution is directly performed based on the graph nodes and their spatial neighbors. This work is mostly inspired by spatial based methods and proposes a graph-based positive prediction (GPP) approach to predict reliable neighbors across graphs formed by different candidates. The proposed GPP is also related to LBFC \cite{wang2019linkage} and SGGNN \cite{shen2018graph}, which are designed for face clustering and re-ID re-ranking, respectively. The main difference is that GPP is designed to explore reliable neighbors that contribute to generalize re-ID networks. Importantly, GPP is trained on hard negative samples selected from the whole dataset, while LBFC and SGGNN are trained with limited samples from a mini-batch.

\section{The Proposed Method}

This paper aims to address the problem of unsupervised domain adaptation (UDA) in person re-identification (re-ID). In the context of UDA in re-ID, we are provided with a fully labeled source domain \{$X_s, Y_s$\} and an unlabeled target domain $X_t$. The source domain includes $N_s$ person images of $M$ identities. Each person image $x^s_i$ is associated with an identity $y^s_i$. The target domain contains $N_t$ person images. The identity annotation of the target domain is not available. In general, the source and target domains are drawn from different distributions, and the identities in the source and target domains are completely different. In this paper, our goal is to learn a transferable deep re-ID model that generalizes well on the target testing set. The labeled source domain and unlabeled target domain are exploited during training.

\subsection{Overview of Framework}

The framework of the proposed method is illustrated in Fig.~\ref{fig:framework}. In our method, the inputs are sampled from the labeled source domain and the unlabeled target domain. The inputs are first fed-forward into the feature extractor to obtain the up-to-date features. The feature extractor is composed of convolutional layers, global average pooling (GAP) \cite{lin2013network} and a batch normalization layer \cite{ioffe2015batch}. The convolutional layers are the residual blocks of ResNet-50 \cite{resnet} pre-trained on ImageNet \cite{deng2009imagenet}. The output of feature extractor is 2,048 dimensional.
Subsequently, the source branch and the target branch are developed for training our model with the source data and the target data, respectively. Each branch includes an exemplar memory module. The memory module is served as a feature-storage that saves the up-to-date output of feature extractor for each source/target sample. The aims of the source branch are twofold. On the one hand, we use the identity classifier with cross-entropy loss to learn basic representations for the feature extractor. On the other hand, we first select the top-ranked features of the input from the source memory. These top-ranked features are regarded as candidate neighbors and are utilized to learn a Graph-based Positive Prediction (GPP) network. The GPP network consists several graph convolution layers \cite{kipf2016semi} and a positive classifier, which aims to predict the probability of a candidate that belongs to a positive sample of the input. The target branch is designed to enforce the proposed three properties of invariance on unlabeled target data, \emph{i.e.} exemplar-, camera- and neighborhood- invariance. Given a target sample, the invariance learning loss is calculated by estimating the similarities between the target sample and whole target samples in the target memory. During invariance learning, we use the learned GPP to infer reliable positive neighbors of the target sample. These neighbors are selected from the candidates of the target memory, according to the probabilities obtained by the positive classifier of GPP. Note that, the GPP network is trained with only the source data. Besides, the loss of the GPP network is not utilized to update the feature extractor.

\subsection{Supervised Learning for Source Domain}

Given the labeled source data, we are able to learn a basic discriminative model in a supervised way. In this paper, we treat the training process of the source domain as an identity classification problem \cite{zheng2016personsurvery}. Thus, the supervised loss of the source domain is calculated by cross-entropy loss, formulated as,
\begin{eqnarray}
    \begin{array}{l}
   \mathcal{L}_{src} = - \log p(y^s_i|x^s_i),
   \label{source-loss}
   \end{array}
\end{eqnarray}
where $p(y^s_i|x^s_i)$ is the predicted probability that the source image $x^s_i$ belongs to identity $y^s_i$. $p(y^s_i|x^s_i)$ is obtained by the identity classifier of the source branch.

It is reported that the model trained using the labeled source data produces a high accuracy on the same distributed testing set. However, without adapting the model, the performance will deteriorate significantly when tested on an unseen (target) testing set. This deterioration is mainly caused by the domain shift and will be more serious with the increase of domain shift. Next, we will introduce the exemplar memory based target invariance learning method to improve the transferability of the model.

\subsection{Invariance Learning for Target Domain}

The deep re-ID model trained with only the source data is usually sensitive to the intra-domain variations of the target domain. In fact, the variations are critical factors influencing the performance in target testing set. Therefore, it is important and necessary to consider the intra-domain variations of the target domain during transferring the knowledge from the source domain to the target domain. To this end, this work investigates three underlying properties of target invariance and adapts the network with the constraints of the three properties. The three properties of target invariance are Exemplar-Invariance (EI), Camera-Invariance (CI) and Neighborhood-Invariance (NI), which are described as follows.

\subsubsection{Three Properties of Target Invariance}

\textbf{Exemplar-invariance.} Given a well-trained re-ID model, the top-ranked results are usually visually similar to the query. This suggests that the model has learned to discriminate persons by apparent similarity. However, this phenomenon may no longer apply when tested on a different distributed dataset. In fact, the appearance of each person image may be very different from others, even if they share the same identity. We call this property as exemplar-invariance, where each person image should be close to itself while far away from others. Therefore, by enforcing the exemplar-invariance on the target images, it is possible to enable the model to improve the ability of distinguishing person images based on the apparent representation.

\begin{figure}[!t]
\centering
\includegraphics[width=\linewidth]{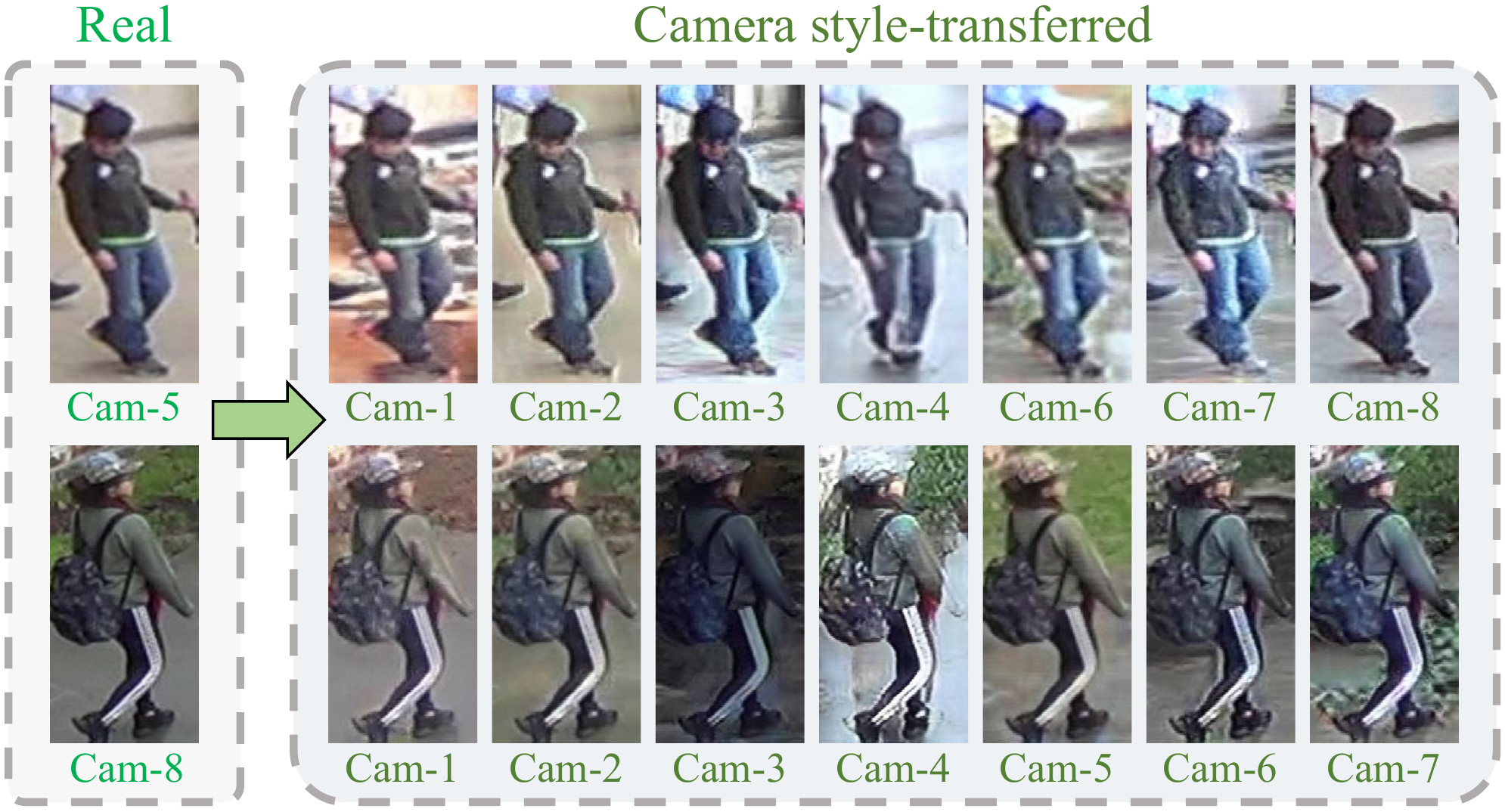}
\caption{Example of camera style-transferred images on DukeMTMC-reID. A real image collected from a certain camera is transferred to images in the styles of other cameras. In this process, the identity information is preserved to some extent. The real image and its corresponding fake images are assumed to belong to the same class during camera-invariance learning.}
\label{fig:camstyle_example}
\end{figure}

\textbf{Camera-invariance.} Camera style variation is a natural and important factor in re-ID, where a person image may encounter significant changes in appearance under different cameras. A model trained using labeled source data can learn the camera-invariance for the source cameras, but may suffer from the variations caused by the target cameras. This is because the camera settings of the two domains will be very different. Inspired by HHL \cite{Zhong_2018_ECCV}, we achieve the property of target camera-invariance by learning with unlabeled target images and their camera style transferred counterparts. These counterparts are in the style of different cameras but share the same identity with the original image. In the constraint of camera-invariance, a target image and its camera-style transferred counterparts are encouraged to be close to each other. Suppose we have $C$ cameras in the target set, we train CamStyle model \cite{zhong2018camera} for the target domain with CycleGAN \cite{zhu2017cyclegan} or StarGAN \cite{stargan}. For Market-1501 \cite{zheng2015scalable} and DukeMTMC-reID \cite{zheng2017unlabeled} that have less cameras, we use CycleGAN to train CamStyle model. For MSMT17 that has many cameras, we apply StarGAN to train CamStyle model. With the learned CamStyle model, each real target image collected from camera $c$ is augmented with $C-1$ images in the styles of other cameras while remaining the original identity. Examples of real images and fake images generated by the CamStyle model are shown in Fig.~\ref{fig:camstyle_example}.

\textbf{Neighborhood-invariance.} Besides the camera-invariance property that is natural and can be easily identified, there are many crucial intra-domain variations that are hard to overcome without the annotation of identity, such as the variations of pose and background. In fact, for each target image, there may exist a number of positive samples in the target dataset. If we could exploit these positive samples during the adapting process, we are able to further improve the robustness of re-ID model in overcoming the variations in the target domain. Assuming that we are given an appropriate transferred model and a query target sample. The nearest neighbors of the query in the target set are most likely to be positive samples of the query. In light of this, we introduce the neighborhood-invariance under the assumption that a target image and its reliable neighbors will share the same identity and should be close to each other.

Next, we will introduce the loss functions of enforcing the three invariance constraints into the network.

\subsubsection{Target Invariance Learning with Memory}

An essential step of enforcing the three properties of invariance is to estimate the similarities (relationships) among target samples. Intuitively, a straightforward solution is to calculate the similarities between samples within a training mini-batch and enforce the three constraints by contrastive loss [18] or triplet loss [22]. However, the number of samples in a training mini-batch is far less than that in the entire training set due to the limited GPU memory. This will cause two problems during invariance learning. First, it is hard to compose an appropriate mini-batch that includes various samples, and their corresponding CamStyle counterparts and neighbors. Second, the similarities between mini-batch samples are local and limited compared to the overall dataset. To address these problems, we introduce an exemplar memory into the network to store the up-to-date representations of the entire target samples. The memory enables the network to directly estimate the similarities between a training target sample and whole target samples, thereby effectively implementing the invariance constraints globally.

\textbf{Exemplar Memory.} The exemplar memory is a feature bank  \cite{xiao2017joint} that stores the up-to-date features of the entire dataset. Given an unlabeled target data including $N_t$ images, we construct an exemplar memory ($\mathcal{F}^t$) which has $N_t$ slots. Each slot stores the L2-normalized output of the feature extractor for the corresponding target image. In the initialization, we initialize the values of all the features in the memory to zeros. During each training iteration, for a target training sample $x^t_i$, we forward it through the model and obtain the L2-normalized  output of the feature extractor, $f(x^t_i)$. 
During the back-propagation, we update the feature in the memory for the training sample $x^t_i$ through,
\begin{eqnarray}
  \begin{array}{l}
   \mathcal{F}^t[i] \leftarrow  \alpha \mathcal{F}^t[i] + (1 - \alpha)  f(x^t_i),
   \label{memory-update}
   \end{array}
\end{eqnarray}
where $\mathcal{F}^t[i]$ is the feature of image $x^t_i$ in the $i$-\emph{th} slot. The hyper-parameter $\alpha$ $\in [0, 1]$ controls the updating rate. $\mathcal{F}^t[i]$ is then L2-normalized via $\mathcal{F}^t[i] \leftarrow \Vert \mathcal{F}^t[i] \Vert_2$. Next, we will introduce the approach of memory based invariance learning.

\textbf{Exemplar-invariance learning.} The exemplar-invariance enforces a target image to be close to itself while far away from others. To achieve this goal, we regard the $N_t$ target images as $N_t$ different classes and apply a non-parameterized manner to classify each image into its own class. For simplicity, we assign the corresponding index as the class of each target sample. Specifically, given a target image $x^t_i$, we first compute the cosine similarities between the embedding $f(x^t_i)$ and features saved in the memory  $\mathcal{F}^t$. Then, the predicted probability that $x^t_i$ belongs to class $i$ is calculated using softmax function,
\begin{eqnarray}
  \begin{array}{l}
  p(i|x^t_i) = \frac{\exp (\mathcal{F}^t[i]^\mathrm{T} f(x^t_i) / \beta)}{\sum_{j=1}^{N_t} \exp (\mathcal{F}^t[j]^\mathrm{T} f(x^t_i) / \beta)},
   \label{eq:probability}
   \end{array}
\end{eqnarray}
where $\beta \in (0, 1]$ is temperature fact that balances the scale of distribution. The objective of exemplar-invariance learning is to minimize the negative log-likelihood over the target image $x^t_i$, as
\begin{eqnarray}
    \begin{array}{l}
   \mathcal{L}_{ei} = - \log p(i|x^t_i).
   \label{exemplar-invariance}
   \end{array}
\end{eqnarray}

\textbf{Camera-invariance learning.} The camera-invariance enforces a target image to be close to its style-transferred counterparts. To introduce the camera-invariance learning into the model, we try to classify each real image and its style-transferred counterparts to the same class. The loss function of camera-invariance is explained as,
\begin{eqnarray}
    \begin{array}{l}
   \mathcal{L}_{ci} = - \log p(i|{\hat{x}}^t_i),
   \label{camera-invariance}
   \end{array}
\end{eqnarray}
where ${\hat{x}}^t_i$ is a target image randomly sampled from the style-transferred images of $x^t_i$. In this way, images in different camera styles of the same sample are forced to be close to each other.

\textbf{Neighborhood-invariance learning.} The neighborhood-invariance enforces a target image to be close to its reliable neighbors. To endow this property into the network, we classify the target image $x^t_i$ into the classes of its reliable neighbors. Supposing the selected reliable neighbors of $x^t_i$ in the memory are denoted as $\mathcal{K}(x^t_i)$. We assign the weight of the probability that $x_{t,i}$ belongs to the class $j$ as,
\begin{eqnarray}
  w_{i, j} =
  \begin{cases}
   \frac{1}{|\mathcal{K}(x^t_i)|}, &  j \neq i  \\ 
    1, & j = i 
  \end{cases}, \forall j \in \mathcal{K}(x^t_i),
\label{weight}
\end{eqnarray}
where $|\mathcal{K}(x^t_i)|$ denotes the size of $\mathcal{K}(x^t_i)$. The objective of neighborhood-invariance learning is formulated as a soft-label loss,
\begin{eqnarray}
    \begin{array}{c}
   \mathcal{L}_{ni} =  - \sum\limits_{j \neq i} w_{i, j} \log p(j|x^t_i), ~~ \forall j\in \mathcal{K}(x^t_i).
   \label{neighborhood-invariance}
   \end{array}
\end{eqnarray}
Note that, in order to distinguish between exemplar-invariance learning and neighborhood-invariance learning, $x^t_i$ is not classified to its own class in Eq.~\ref{neighborhood-invariance}.

The key of neighborhood-invariance learning is including as many positive samples as possible in $\mathcal{K}(x^t_i)$ while rejecting negative samples.  In Section \ref{Section:GPP}, we will introduce a vanilla neighbor selection (VNS) method and a graph-based positive prediction (GPP) method for exploring reliable neighbors $\mathcal{K}(x^t_i)$.

\begin{figure*}[!t]
    \centering
    \includegraphics[width=0.98\linewidth]{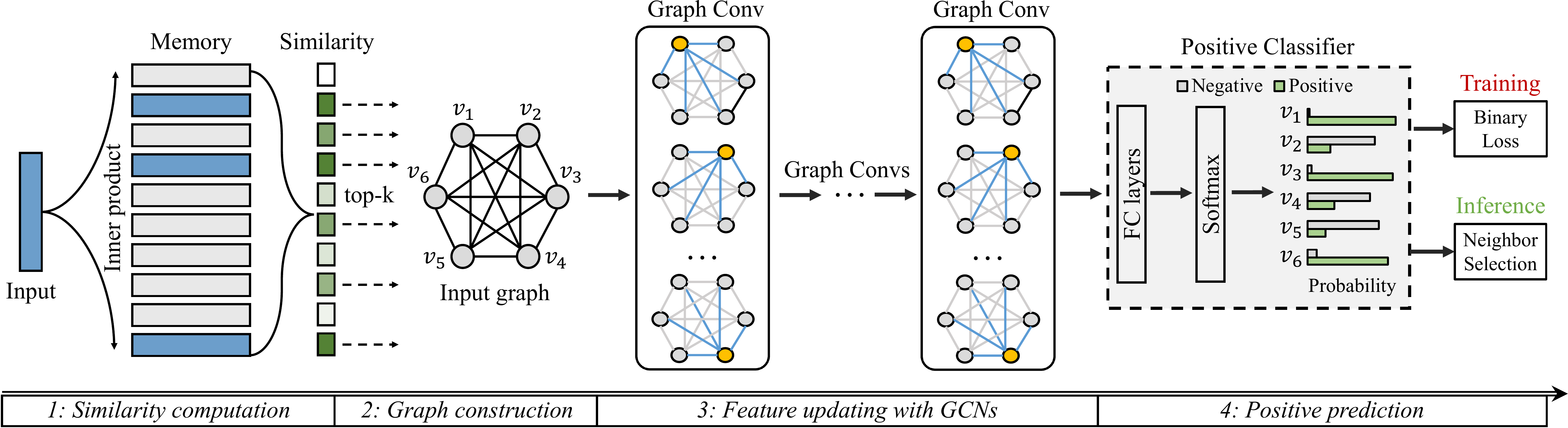}
    \caption{The pipeline of graph-based positive prediction (GPP). Given the embedding of an input sample, 1) we first compute the similarities between the input and features in the memory. 2) The top-k ranked samples are selected as candidate neighbors and utilized to construct a graph for positive prediction. 3) The features of nodes are refined by graph convolutional networks (GCNs) on the graph. 4) The positive classifier is employed to predict positive probabilities of every node. For the source domain, the positive classification loss is computed for training the network of GPP. For the target domain, we select reliable neighbors for the input target sample, depending on the predicted positive probabilities of nodes.}
    \label{fig:pipeline-gcn}
\end{figure*}

\textbf{Overall loss of invariance learning.} By jointly considering the exemplar-invariance, camera-invariance and neighborhood-invariance, the overall loss of invariance learning over a target training image can be written as,
\begin{eqnarray}
    \begin{array}{l}
   \mathcal{L}_{tgt} =  - \sum\limits_{j} w_{i, j} \log p(j|\widetilde{x}^t_i), j \in \mathcal{K}(\widetilde{x}^t_i),
   \label{target-loss}
   \end{array}
\end{eqnarray}
where $\widetilde{x}^t_i$ is an image randomly sampled from the union set of $x^t_i$ and its camera style-transferred images. In Eq.~\ref{target-loss}, when $i=j$, we optimize the network with the exemplar-invariance learning and camera-invariance learning by classifying $\widetilde{x}^t_i$ into its own class. When $i \neq j$, the network is optimized with the neighborhood-invariance learning by leading $\widetilde{x}^t_i$ to be close to its reliable neighbors in $\mathcal{K}(\widetilde{x}^t_i)$.

\subsection{Graph-based Positive Prediction}
\label{Section:GPP}

\textbf{Vanilla Neighbor Selection (VNS).} A naive way to select neighbors of a target image $x^t_i$ is based on the cosine similarities between $f(x^t_i)$ and features saved in the memory. In the vanilla neighbor selection (VNS) method, we directly select the top-$k$ ranked samples from the memory as the $\mathcal{K}(x^t_i)$. However, in such simple approach, the similarities between $f(x^t_i)$ and features are estimated independently, neglecting the relationships among the features in the memory. As a consequence, the similarity estimation of hard positive and negative samples may not be accurate, especially when the model has poor performance on the target domain.

To overcome this problem, next, we will propose a novel neighbor selection method, called graph-based positive prediction (GPP). The network of GPP is constructed by graph convolutional networks (GCNs) \cite{kipf2016semi} and positive classifier. The goal of GPP is to refine the similarities between $f(x^t_i)$ and features in the memory by leveraging the relationships among the features. To achieve this goal, the network of GPP is trained to predict probabilities of samples in the memory that belong to positive class of the input sample. We train the GPP network with labeled source samples and then apply it to infer reliable neighbors for target training samples.

\subsubsection{Training GPP on Source Domain}

To train the network of GPP, we simulate the positive prediction process on the source domain. In the implementation, we construct a source memory $\mathcal{F}^s$ with $N_s$ slots for storing the features of source samples. The network of GPP includes several graph convolutional layers and a positive classifier. The training process of GPP is illustrated in Fig.~\ref{fig:pipeline-gcn}, which is divided into four steps as described below.

\textbf{1. Similarity computation.} Given a training source sample $x^s_i$, we first extract its feature $f(x^s_i)$, and compute the cosine similarities between $f(x^s_i)$ and features in the source memory $\mathcal{F}^s$.

\textbf{2. Graph construction.} We select $k$-nearest-neighbors from the ranked-list as candidate neighbors of $x^s_i$, which is denoted as $V=\{v_1, v_2, ..., v_k\}$. Then, we construct a complete undirected graph $G(V, E)$, where $V$ denotes the set of nodes and $E$ indicates the set of edges. We denote $\mathcal{H}=\{\mathcal{F}^s[v_1], \mathcal{F}^s[v_2], ..., \mathcal{F}^s[v_k]\}$ as the node features. In order to encode the information of $x^s_i$, we normalize $\mathcal{H}$ by subtracting $f(x^s_i)$,
\begin{eqnarray}
    \begin{array}{l}
   \mathcal{H}=\{\mathcal{F}^s[v_1]-f(x^s_i), ..., \mathcal{F}^s[v_k]-f(x^s_i)\}.
   \label{norm-F}
   \end{array}
\end{eqnarray}
%
In practice, the $\mathcal{H}$ can be represented by a matrix with a size of $k \times d$, where $d$ is the feature dimension of each node. As well, the weights of $E$ are represented by an adjacency matrix $\mathcal{A} \in \mathbb{R}^{k \times k}$, where
\begin{eqnarray}
    \begin{array}{l}
   \mathcal{A}_{i, j} = \mathcal{H}_i^{T} \mathcal{H}_j, ~~\forall i, j \in V.
   \label{matrix}
   \end{array}
\end{eqnarray}
Lastly, each row of $\mathcal{A}$ is normalized by softmax function.

\textbf{3. Feature updating with GCNs}. In this step, we aim to improve the node representations with the neighbor relations, so that can accurately predict positive samples from the candidates. To achieve this goal, we employ the graph convolution network (GCNs) \cite{kipf2016semi,wang2019linkage} to update node features on $G(V, E)$. The input to the GCNs is a set of node features $\mathcal{H}$ and an adjacency matrix $\mathcal{A}$, and the output is a new set of node features $\mathcal{Z}$. Every graph convolutional layer in the GCNs can be written as a non-linear function,
\begin{eqnarray}
    \begin{array}{l}
   \mathcal{H}^{(l+1)} = \rm{ReLU}([\mathcal{A}\mathcal{H}^{(\textit{l})} || \mathcal{H}^{(\textit{l})}]\mathcal{W}^{(\textit{l})}),
   \label{gcn-opereation}
   \end{array}
\end{eqnarray}
with $\mathcal{H}^{(0)}= \mathcal{H}$ and $\mathcal{Z}=\mathcal{H}^{(L)}$. $L$ is the number of graph convolutional layers. $||$ is the matrix concatenation operation. $\mathcal{W}^{(l)} \in \mathbb{R}^{2 d_{in} \times d_{out}}$ is a learnable weight matrix for the $l$-th graph convolutional layer. $d_{in}$ and $d_{out}$ are the dimensions of input feature and out feature, respectively. In this paper, we adopt 4 graph convolutional layers to form the GCNs. The output $\mathcal{Z}$ is used to predict the positive probabilities for nodes by a positive classifier, as described in the next step.

\begin{figure*}[!t]
    \centering
    \includegraphics[width=\linewidth]{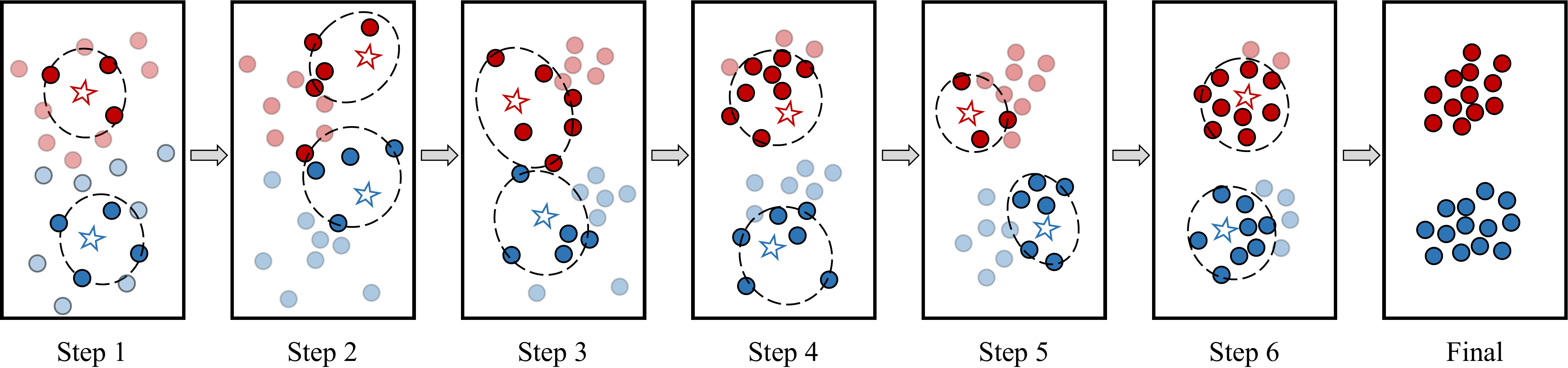}
    \caption{Toy example of invariance learning. Dot colors denote classes. In each step, an input and its reliable neighbors (highlighted in circle) are enforced to be close by neighborhood-invariance learning, while an input and other samples (out of circle) are enforced to be far away by exemplar-invariance learning. With the interaction of exemplar-invariance and neighborhood-invariance, samples with the same class are gradually grouped closer, while dissimilar groups are separated.}
    \label{fig:toy-example}
\end{figure*}

\textbf{4. Prediction with positive classifier}. Given the updated features $\mathcal{Z}$, we use a positive classifier to predict the probability that a node belongs to the positive sample of the input $x^s_i$. For training of the GPP network, the loss function on the source domain is formulated by,
\begin{eqnarray}
    \begin{array}{l}
   \mathcal{L}_{gpp} = - \frac{1}{k} \sum\limits_{j} (y^*_j \log p^*_j + (1 - y^*_j) \log (1 - p^*_j)),
   \label{orr-loss}
   \end{array}
\end{eqnarray}
where $j \in V$. $p^*_j$ is the predicted probability that node $j$ belongs to the positive of the input $x^s_i$. $y^*_i$ is the ground-truth binary label, defined as,
\begin{eqnarray}
  y^*_j =
  \begin{cases}
   0, & y^s_j \neq y^s_{i} \\ 
    1, & y^s_j = y^s_{i} 
  \end{cases}, \forall j \in V.
\label{binary-label}
\end{eqnarray}
We optimize the network of GPP with $\mathcal{L}_{gpp}$ computed on the labeled source data.

\subsubsection{Reliable Neighbors Selection on Target Domain}

We infer the network of GPP on the target domain with the same steps as the training process on the source domain, except the loss computation in the last step. For a target sample $x^t_i$, we first compute the similarities between $f(x^t_i)$ and features in the target memory $\mathcal{F}^s$ with step 1. Then, we obtain the positive probabilities for the candidate neighbors $V$ with the remaining steps. Finally, the reliable neighbors $\mathcal{K}(x^t_i)$ are selected depending on the positive probabilities, defined as,
\begin{eqnarray}
  \mathcal{K}(x^t_i)=\{j|j \in V \land p^*_j \geq  \mu \},
\label{eq:reliable-neighbor}
\end{eqnarray}
where $\mu$ is threshold value that controls whether a candidate should be selected as a reliable neighbor.

\subsection{Final Loss for Network}

By combining the losses of supervised learning, invariance learning and graph-based positive prediction learning, the final loss for the overall network is formulated as,
\begin{eqnarray}
    \begin{array}{l}
   \mathcal{L} =  \mathcal{L}_{src} + \mathcal{L}_{tgt} + \mathcal{L}_{gpp}.
   \label{eq:final-loss}
   \end{array}
\end{eqnarray}
%
To this end, we introduce a framework for UDA in person re-ID, in which $\mathcal{L}_{src}$ aims to maintain a basic representation for person. Meanwhile, $\mathcal{L}_{tgt}$ attempts to take the knowledge from the labeled source domain and incorporate the invariance properties of the target domain into the network. $L_{gpp}$ tries to learn a reliable neighbor prediction network that could facilitate the invariance learning. Note that, $L_{gpp}$ is only calculated on the source domain and is only used to update the network of GPP.

\subsection{Discussion on the Three Invariance Properties}
We analyze the advantage and disadvantage for the three invariance properties, and the mutual effectiveness among them. The exemplar-invariance enforces each exemplar away from each other. 
It is beneficial to enlarge the distance between exemplars from different identities. 
However, exemplars of the same identity will also be far apart, which is harmful to the system. 
On the contrast, neighborhood-invariance encourages each exemplar and its neighbors to be close to each other. 
It is beneficial to reduce the distance between exemplars of the same identity. 
However, neighborhood-invariance might also pull closer images of different identities, because we could not guarantee that each neighbor shares the same identity with the input exemplar. Therefore, there exists a trade off between exemplar-invariance and neighborhood-invariance, where the former aims to lead the exemplars from different identities to be far away while the latter attempts to encourage exemplars of the same identity to be close to each other.
In other words, the interaction process between exemplar-invariance and neighborhood-invariance can be considered as a kind of local-clustering. On the one hand, when enforcing neighbor-invariance, samples with the same identity would be progressively grouped closer, through the neighborhood relation and the connection of their shared neighbors. On the other hand, exemplar-invariance encourages dissimilar samples to be pushed away from each other, so that dissimilar groups would be separated. Camera-invariance has the similar effect as the exemplar-invariance and also leads the exemplar and its camera-style transferred samples to share the same representation. A toy example of invariance learning is shown in Fig.~\ref{fig:toy-example}.

\section{Experiment}

\subsection{Dataset} We evaluate the proposed method on three large-scale person re-identification (re-ID) benchmarks: Market-1501 \cite{zheng2015scalable}, DukeMTMC-reID \cite{zheng2017unlabeled} and MSMT17 \cite{wei2018person}. 

\textbf{Market-1501} \cite{zheng2015scalable} includes 32,668 labeled person images of 1,501 identities collected from six non-overlapping camera views. For evaluation, the dataset is divided into 12,936 images of 751 identities for training, 3,368 query images and 19,732 images of 705 identities for testing. 

\textbf{DukeMTMC-reID} \cite{zheng2017unlabeled} is a re-ID benchmark collected from the DukeMTMC dataset \cite{ristani2016performance}. The dataset is captured from eight cameras, including 36,411 person images from 1,812 identities. It contains 16,522 images of 702 identities for training, 2,228 query images of 702 identities and 17,611 gallery images for testing. 

\textbf{MSMT17} \cite{wei2018person} is a newly released large-scale person re-ID benchmark. It is composed of 126,411 person images from 4,101 identities collected by an 15-camera system. The training set consists of 32,621 images of 1,041 identities, and the testing set contains 11,659 images as query and 82,161 images as gallery. The dataset has serious variations of scene and lighting, and is more challenging than the other two benchmarks. 

\textbf{Evaluation Protocol.} During training, we use a labeled training dataset as the source domain and an unlabeled training dataset as the target domain. In testing, performance is evaluated on the target testing set by the cumulative matching characteristic (CMC) and mean Average Precision (mAP). For CMC, we use rank-1, rank-5, rank-10, and rank-20 as metrics.

\subsection{Experimental Settings}

\textbf{Deep re-ID model.} We adopt ResNet-50 \cite{resnet} as the backbone of the feature extractor and initialize the model with the parameters pre-trained on ImageNet \cite{deng2009imagenet}. Specifically, we fix the first two residual layers and set the stride size of last residual layer to 1. After the global average pooling (GAP) layer, we add a batch normalization (BN) layer \cite{ioffe2015batch} followed by ReLU \cite{nair2010relu} and Dropout \cite{srivastava2014dropout}. The identity classifier is an $M$-dimensional FC layer followed by softmax function. The input image is resized to 256 $\times 128$. During training, we perform random flipping, random cropping and random erasing \cite{zhong2017random} for data augmentation. The probability of dropout is set to 0.5. We train the re-ID model with a learning rate of 0.01 for ResNet-50 base layers and of 0.1 for the others in the first 40 epochs. The learning rate is divided by 10 for the next 20 epochs. The SGD optimizer is used to train the re-ID model with a mini-batch size of 128 for both source and target domains. We initialize the updating rate of memory $\alpha$ to 0.01 and increase $\alpha$ linearly with the number of epochs, \emph{i.e.}, $\alpha=0.01 \times epoch$. Without specification, we set the temperature fact $\beta = 0.05$. For vanilla neighbor selection, we set $k=8$ and directly select the top-$k$ nearest-neighbors from the memory as the reliable neighbors. For GPP, we set the number of candidate neighbors $k=100$ and neighbor selection threshold $\mu = 0.9$. We train the model with exemplar-invariance and camera-invariance learning at the first 10 epochs and add the neighborhood-invariance learning for the rest epochs. In testing, we extract the L2-normalized output of GAP as the image feature and adopt the Euclidean metric to measure the similarities between query and gallery images.

\textbf{Network of graph-based positive prediction (GPP).} The GPP network contains four graph convolutional layers and a positive classifier.
The input and out dimensions of these graph convolutional layers are: 2048$\rightarrow$2048, 2048$\rightarrow$512, 512$\rightarrow$256, 256$\rightarrow$256. The positive classifier is composed of a 256-dimensional FC layer, a BN layer, a PReLU layer \cite{he2015PReLU}, and a 2-dimensional FC layer. We utilize SGD optimizer to update the GPP network after the 5\textit{th} epoch. The learning rate is initialized as 0.01 and divided by 10 after the 40\textit{th} epoch.

\textbf{Baseline setting.} We set the model as the \emph{baseline} when trained the network using only the identity classifier. We employ the baseline in two ways: 1) \textit{Train on target}, training the baseline on
the labeled target training data and testing on the target testing set, and 2) \textit{Source only}, training the baseline on a source labeled training set and directly applying to a target testing set without modification. The ``Train on target'' and the ``Source only'' can be considered as the upper bond and lower bond of our method, respectively.

\begin{table}[!t]
\caption{\label{tabel:temperature} Evaluation with different values of $\beta$ in Eq.~\ref{eq:probability}.}
\vspace{-.2in}
\begin{center}
\newcolumntype{C}{>{\centering\arraybackslash}X}%
\newcolumntype{R}{>{\raggedleft\arraybackslash}X}%
\begin{tabularx}{\linewidth}{ c||CC||CC}
\hline
\multicolumn{1}{c||}{\multirow{2}{*}{$\beta$}} & \multicolumn{2}{c||}{Duke $\rightarrow$ Market} & \multicolumn{2}{c}{Market $\rightarrow$ Duke} \\ 
\cline{2-5}
& Rank-1&mAP&Rank-1&mAP\\
\hline
\hline
0.01 & 50.1 & 21.1&45.6	&21.5\\
0.03 & 79.4 & 54.5& 68.1&45.6\\
0.05 & \textbf{84.1} & \textbf{63.8}&\textbf{74.0}&\textbf{54.4}\\
0.1 & 75.7 & 46.5&63.0&39.8\\
0.5 & 59.0 & 30.9&51.0&28.0\\
1.0 & 57.9 & 30.3&44.6&23.8\\
\hline
\end{tabularx}
\end{center}
\end{table}

\begin{figure}[!t]
\centering
\includegraphics[width=\linewidth]{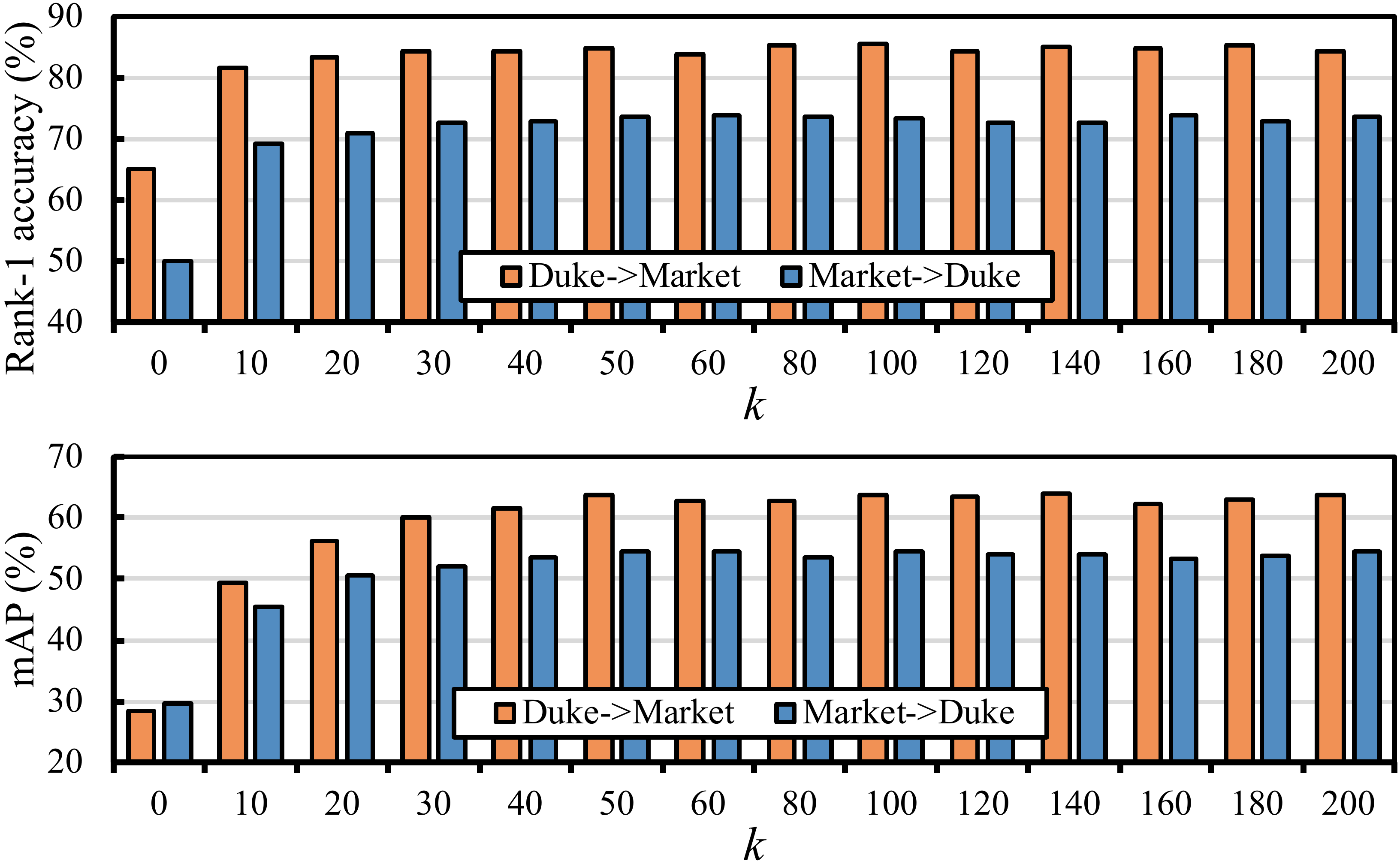}
\caption{Evaluation with different number of candidate samples for graph-based positive prediction.}
\label{fig:different_k}
\end{figure}

\begin{table*}[!t]
\caption{\label{tabel:ablation} Methods comparison when transferred to Market-1501 and DukeMTMC-reID. \textbf{Train on target}: baseline model trained with the labeled target training data. \textbf{Source Only}: baseline model trained with only labeled source data. \textbf{EI}: exemplar-invariance. \textbf{CI}: camera-invariance. \textbf{NI}: neighborhood-invariance.}
\vspace{-.2in}
\begin{center}
\newcolumntype{C}{>{\centering\arraybackslash}X}%
\newcolumntype{R}{>{\raggedleft\arraybackslash}X}%
\begin{tabularx}{\linewidth}{ l|ccc||CCCCC||CCCCC}
\hline
\multicolumn{1}{l|}{\multirow{2}{*}{Methods}}&\multicolumn{3}{c||}{\multirow{1}{*}{ Invariance}}&\multicolumn{5}{c||}{DukeMTMC-reID $\rightarrow$ Market-1501}&\multicolumn{5}{c}{Market-1501 $\rightarrow$ DukeMTMC-reID}\\
\cline{2-14}
\multicolumn{1}{c|}{}&\multicolumn{1}{c}{EI}&\multicolumn{1}{c}{CI}&\multicolumn{1}{c||}{NI}&R-1&R-5&R-10&R-20&mAP&R-1&R-5&R-10&R-20&mAP\\
\hline
\hline
Train on target&-&-&-&87.6& 95.5& 97.2& 98.3&69.4&75.6& 87.3& 90.6& 92.9&57.8\\
\hline
Source only &-&-&-& 43.1& 58.8& 67.3& 74.3& 17.7&28.9&44.0&50.9&57.5&14.8\\
\hline
Ours &\checkmark&-&-&48.7&67.4&74.0&80.2&21.0&34.2&51.3&58&64.2&18.7\\
Ours &\checkmark&\checkmark&-&63.1&79.1&84.6&89.1&28.4&53.9&70.8&76.1&80.7&29.7\\
Ours &\checkmark&-&\checkmark&71.8&83.1&87.1&90.6&45.7&67.2&80.0&83.8&86.7&48.3\\
Ours &\checkmark&\checkmark&\checkmark&\bf 84.1&\bf 92.8&\bf 95.4&\bf 96.9&\bf 63.8&\bf74.0&\bf 83.7&\bf 87.4&\bf 90.0&\bf 54.4\\
\hline
\end{tabularx}
\end{center}
\end{table*}

\subsection{Parameter Analysis}

We first investigate the sensitivities of our approach to three important hyper-parameters, \emph{i.e.}, the temperature fact $\beta$ in Eq.~\ref{eq:probability}, the number of candidate neighbors $k$ in graph $G(V, E)$ and the neighbor selection threshold $\mu$ in Eq.~\ref{eq:reliable-neighbor}. In order to clearly analyze the impact of every parameter, we vary the value of one parameter and keep fixed the others. Experiments are evaluated on the setting of transferring between Market-1501 and DukeMTMC-reID.

\textbf{Temperature fact $\beta$}. In Table~\ref{tabel:temperature}, we report the impact of the temperature fact $\beta$ in Eq.~\ref{eq:probability}. Assigning a lower value to $\beta$ gives rise to a lower entropy, which commonly produces better results. However, when assigning a extremely low value to $\beta$, the model does not converge and produces a poor performance, \emph{e.g.}, $\beta = 0.01$. The best results are obtained at $\beta=0.05$.

\textbf{Number of candidate neighbors $k$}. In Fig.~\ref{fig:different_k}, we evaluate the performance of using a different number of candidate neighbors in the graph $G(V, E)$. When $k=0$, our approach reduces to the model trained with only exemplar-invariance and camera-invariance. It is evident that injecting neighborhood-invariance into the network ($k>0$) can significantly improve accuracy, and our method is insensitive to the changes of $k$. The rank-1 accuracy and mAP first improve with the increase of $k$ and reach stable when $k>100$. This improvement tendency is reasonable, because: 1) Using a larger $k$ will include more positive samples in the graph and GPP might discover more hard positive samples for neighborhood-invariance learning; 2) Most positive samples are within the top-100 nearest neighbours, so it is unnecessary to include too many candidate samples. Considering the trade-off between accuracy and speed, we set $k=100$ in our approach.

\begin{figure}[!t]
\centering
\includegraphics[width=\linewidth]{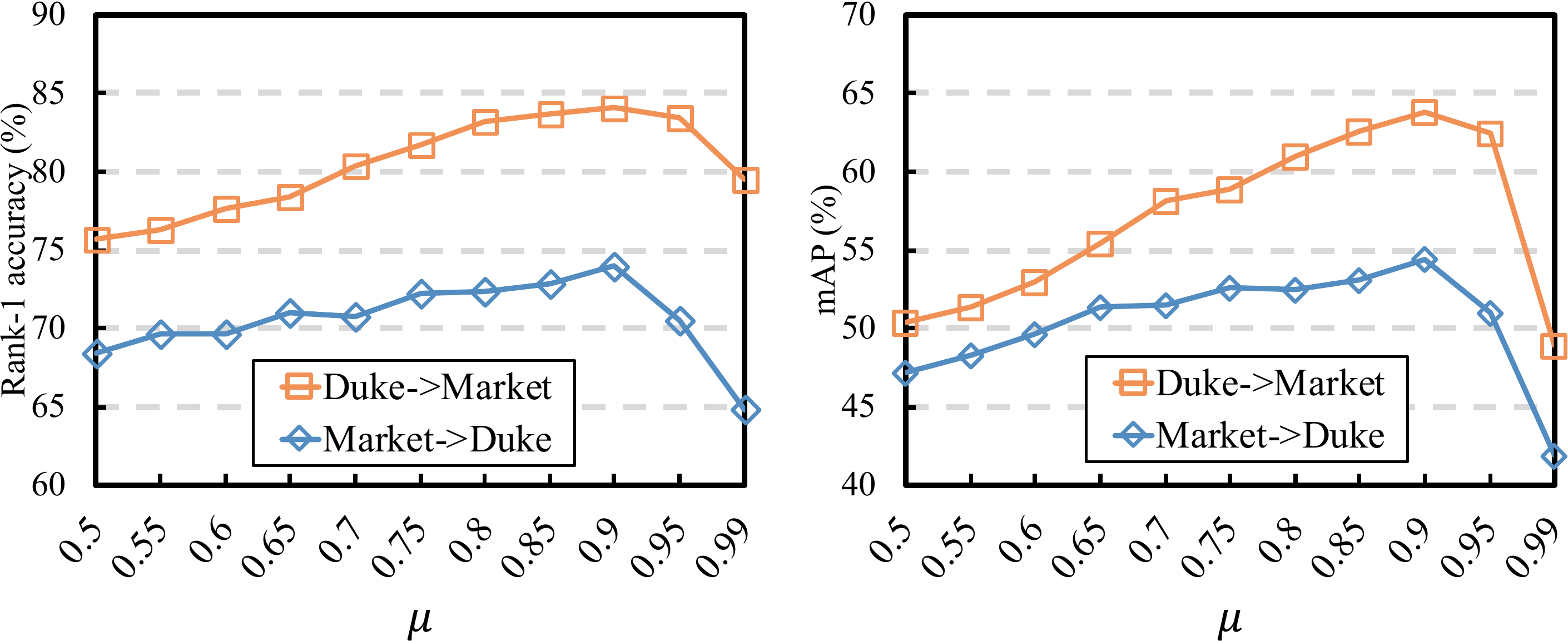}
\caption{Evaluation with different values of $\mu$ in Eq.~\ref{eq:reliable-neighbor}}
\label{fig:mu}
\end{figure}

\textbf{Threshold of positive neighbor selection $\mu$.} In Fig.~\ref{fig:mu}, we compare the performance of using different values of $\mu$ in Eq.~\ref{eq:reliable-neighbor}. On the one hand, assigning a too high value to $\mu$ would only select easy positive samples for neighbor-invariance learning, \textit{e.g.}, $\mu=0.99$. This will result in the model suffers from hard positive samples in testing. Note that, when $\mu=1$, our method reduces to the model trained without neighbor-invariance learning. Since extremely few samples would be selected as reliable positive neighbors. On the other hand, giving a too low value to $\mu$ might include too many false positive samples for neighbor-invariance learning, \textit{e.g.}, $\mu<0.7$. Approaching a sample to false positive samples would undoubtedly have deleterious effects on the results. Our approach reaches the best result when $\mu$ is around 0.9.

According to above analysis, we set $\beta = 0.05$, $k=100$ and $\mu=0.9$ in the following experiments.

\subsection{Evaluation and Analysis}

\textbf{Performance of baseline in domain adaptation.} In Table~\ref{tabel:ablation}, we report the results of the baseline when transferring between Market-1501 and DukeMTMC-reID. When trained on the labeled target training set, the baseline (\emph{Train on target}) achieves high accuracy. However, we observe a serious drop in performance when the baseline (\emph{Source only}) is trained only using the labeled source set  and directly applied to the target testing set. For the case of testing on Market-1501, the baseline (\emph{Train on target}) achieves a rank-1 accuracy of 87.6\%, but the rank-1 accuracy of the baseline (\emph{Source only}) declines to 43.1\%. A similar drop can be observed when transferred from Market-1501 to DukeMTMC-reID. This decline in accuracy is mainly caused by the domain shifts between datasets.

\textbf{Ablation study on invariance learning.} To study the effectiveness of the proposed invariance learning of target domain, we conduct ablation experiments in Table~\ref{tabel:ablation}. We start from the basic model (Our method w/ EI), which enforces exemplar-invariance learning into the baseline (\emph{Source only}) model, and then add camera-invariance, neighborhood-invariance, and both.

First, we show the effect of adding exemplar-invariance learning. As shown in Table~\ref{tabel:ablation}, ``Ours w/ EI'' consistently improves the results over the baseline (\emph{Source only}). Specifically, the rank-1 accuracy improves from 43.1\% to 48.7\% and 28.9\% to 34.2\% when tested on Market-1501 and DukeMTMC-reID, respectively. This demonstrates that exemplar-invariance learning is an effective way to improve the discrimination of person descriptors for the target domain.

\begin{table}[!t]
\caption{\label{tabel:number of camstyle} Evaluation with different number of camera style samples for each target image. $C$ is the number of cameras in the target domain. Model is trained with all three invariance constraints.}
\vspace{-.2in}
\begin{center}
\newcolumntype{C}{>{\centering\arraybackslash}X}%
\newcolumntype{R}{>{\raggedleft\arraybackslash}X}%
\begin{tabularx}{\linewidth}{ c||CC||CC}
\hline
\multicolumn{1}{c||}{\multirow{1}{*}{\# CamStyle}} & \multicolumn{2}{c||}{Duke $\rightarrow$ Market} & \multicolumn{2}{c}{Market $\rightarrow$ Duke} \\ 
\cline{2-5}
samples & Rank-1&mAP&Rank-1&mAP\\
\hline
\hline
0 &71.8&45.7&67.2&48.3\\
1 & 83.5&61.1&72.1&52.3\\
3 &84.0&62.8&73.0&53.0\\
$C$-1 & \textbf{84.1} & \textbf{63.8}&\textbf{74.0}&\textbf{54.4}\\
\hline
\end{tabularx}
\end{center}
\end{table}

\begin{table*}[!t]
\caption{\label{tabel:memory and GPP} Results and computational cost analysis of the exemplar memory and graph-based positive prediction (GPP).}
\vspace{-.2in}
\begin{center}
\newcolumntype{C}{>{\centering\arraybackslash}X}%
\newcolumntype{R}{>{\raggedleft\arraybackslash}X}%
\begin{tabularx}{0.75\linewidth}{cc|c|c||CC||CC}
\hline
\multicolumn{2}{c|}{Module}&Training time&GPU memory&\multicolumn{2}{c||}{Duke$\rightarrow$ Market} & \multicolumn{2}{c}{Market$\rightarrow$ Duke}\\
\cline{1-2}
\cline{5-8}
Memory&GPP&(s/iter)&(MB)&R-1 & mAP &R-1 & mAP\\
\hline
\hline
-&-&$\approx$0.50&$\approx$6,800&70.2&40.6&55.3&33.7\\
\checkmark&-& $\approx$0.52&$\approx$7,000&77.4&45.5&65.4&42.6\\
\checkmark&\checkmark&$\approx$0.63&$\approx$9,800&84.1&63.8&74.0&54.4\\
\hline
\end{tabularx}
\end{center}
\end{table*}

\begin{table}[!t]
\caption{\label{tabel:neighor selection} Results of using different neighbor selection methods. \textbf{VNS:} Vanilla neighbor selection. \textbf{GPP:} graph-based positive prediction.}
\vspace{-.2in}
\begin{center}
\newcolumntype{C}{>{\centering\arraybackslash}X}%
\newcolumntype{R}{>{\raggedleft\arraybackslash}X}%
\begin{tabularx}{\linewidth}{l|l|CC||CC}
\hline
\multicolumn{1}{l|}{\multirow{1}{*}{Neighbor}}&\multicolumn{1}{l|}{\multirow{2}{*}{Condition}}&\multicolumn{2}{c||}{Duke$\rightarrow$ Market}&\multicolumn{2}{c}{Market$\rightarrow$Duke}\\
\cline{3-6}
selection&&R-1 & mAP &R-1 & mAP\\
\hline
\hline
VNS&top-8&77.4&45.5&65.4&42.6\\
Variant VNS &$p\geq0.9$&77.0&46.4&64.5&42.8\\
Variant GPP &top-8&79.8&51.3&68.2&45.3\\
GPP&$p\geq0.9$&\bf 84.1&\bf 63.8&\bf 74.0&\bf 54.4\\
\hline
\end{tabularx}
\end{center}
\end{table}

Second, we validate the effectiveness of camera-invariance learning over the basic model (ours w/ EI). In Table~\ref{tabel:ablation}, we observe significant improvement when adding camera-invariance learning into the system. For example, ``Ours w/ EI+CI'' achieves a rank-1 accuracy of 63.1\% when transferred from DukeMTMC-reID to Market-1501. This is higher than ``Ours w/ EI'' by 14.4\% in rank-1 accuracy. The improvement demonstrates that the image variations caused by target cameras severely impact the performance in target testing set. Injecting camera-invariance learning into the model could effectively improve the robustness of the system to camera style variations. In Table~\ref{tabel:number of camstyle}, we also report the results of using different number of camera style samples for each target image. Even with only one camera style sample for each target image, the results of our approach can be considerable improved and are slightly lower than that of using all camera style samples. This indicates that the number of camera style samples in our approach can be agnostic to the number of cameras, and thus our approach is scalable to a scenario with many cameras.

Third, we evaluate the effect of neighborhood-invariance learning. As reported in Table~\ref{tabel:ablation}, neighborhood-invariance significantly improves performance of the basic model (Ours w/ EI). When adding neighborhood-invariance to the basic model, ``Ours w/ EI+NI'' obtains 67.2\% rank-1 accuracy and 48.3\% mAP when transferred from Market-1501 to DukeMTMC-reID. This increases the results of the basic model by 33\% in rank-1 accuracy and by 29.7\% in mAP, respectively. A similar improvement is achieved when transferred to Market-1501.

Finally, we demonstrate the mutual benefit among the three invariance properties. As shown in the last row in Table~\ref{tabel:ablation}, the three invariance properties are complementary to each other. The integration of them achieves obvious improvement over independently adding camera-invariance or neighborhood-invariance to the basic model. For example, ``Ours w/ EI+CI+NI'' achieves rank-1 accuracy of 84.1\% when transferred from DukeMTMC-reID to Market-1501, outperforming ``Ours w/ EI+CI'' by 21\% and ``Ours w/ EI+NI'' by 12.3\%. Similar improvement is observed when transferred to DukeMTMC-reID. Particularly, our final model (Ours w/ EI+CI+NI) has only a small gap with the upper bond model (\emph{Train on target}). For instance, our final model reaches rank-1 accuracy of 74.0\% and mAP of 54.4\% when transferred to DukeMTMC-reID, which is lower than ``Train on target'' by 1.6\% in rank-1 accuracy and by 3.4\% in mAP. This demonstrates that our approach has a strong capacity of bridging the gap between person re-ID domains.

\begin{table*}[!t]
\caption{\label{tabel:sota-m-d} Unsupervised person re-ID performance comparison with state-of-the-art methods on Market-1501 and DukeMTMC-reID. \textbf{Market}: Market-1501 \cite{zheng2015scalable}. \textbf{Duke}: DukeMTMC-reID \cite{zheng2017unlabeled}. \textbf{MSMT}: MSMT17 \cite{wei2018person}. \textbf{SyRI}: SyRI \cite{Bak_2018_ECCV}. \textbf{Multi}: a combination of seven datasets.}
\vspace{-.2in}
\begin{center}
\newcolumntype{C}{>{\centering\arraybackslash}X}%
\newcolumntype{R}{>{\raggedleft\arraybackslash}X}%
\begin{tabularx}{\linewidth}{ l||c||C|CCCC||C|CCCC}
\hline
\multicolumn{1}{l||}{\multirow{2}{*}{Methods}}&\multicolumn{1}{l||}{\multirow{2}{*}{Reference}} &\multicolumn{5}{c||}{Market-1501}&\multicolumn{5}{c}{DukeMTMC-reID}\\
\cline{3-12}
\multicolumn{1}{c||}{}&&Source&R-1&R-5&R-10&mAP&Source&R-1&R-5&R-10&mAP\\
\hline
\hline
LOMO \cite{liao2015lomo}&CVPR 15&-&27.2&41.6&49.1&8.0&-&12.3&21.3&26.6&4.8\\
BOW \cite{zheng2015scalable}&ICCV 15&-&35.8&52.4&60.3&14.8&-&17.1&28.8&34.9&8.3\\
\hline
UMDL \cite{peng2016unsupervised}&CVPR 16&Duke&34.5&52.6&59.6&12.4&Market&18.5&31.4&37.6&7.3\\
PTGAN \cite{wei2018person}&CVPR 18&Duke&38.6 & - & 66.1 & - &Market& 27.4 & - & 50.7 & -\\
PUL \cite{fan2017pul}&TOMM 18&Duke&45.5&60.7&66.7&20.5&Market&30.0&43.4&48.5&16.4\\
SPGAN \cite{deng2018image}&CVPR 18&Duke& 51.5&70.1&76.8& 22.8&Market&41.1&56.6&63.0&22.3\\
CAMEL \cite{yu2017cross}&ICCV 17&Multi&54.5&73.1&-&26.3&Multi&40.3&57.6&-&19.8\\
MMFA \cite{lin2018multibmvc}&BMVC 18&Duke& 56.7&75.0&81.8& 27.4&Market&45.3& 59.8& 66.3& 24.7\\
{SPGAN+LMP} \cite{deng2018image}&CVPR 18&Duke& 57.7&75.8&82.4&26.7&Market& 46.4&62.3&68.0&26.2\\
TJ-AIDL \cite{wang2018reid}&CVPR 18&Duke&58.2&74.8&81.1& 26.5&Market& 44.3& 59.6& 65.0& 23.0\\
CamStyle \cite{zhong2019camstyle}&TIP 19&Duke&58.8& 78.2& 84.3&27.4&Market&48.4&62.5& 68.9&25.1\\
DECAMEL \cite{yu2018unsupervised} &TPAMI 19&Multi&60.2&76.0&-&32.4&-&-&-&-&-\\
HHL \cite{Zhong_2018_ECCV}&ECCV 18&Duke&62.2&78.8&84.0&31.4&Market& 46.9&61.0&66.7&27.2\\
DASy \cite{Bak_2018_ECCV} &ECCV 18&SyRI& 65.7&-&-&-&-&-&-&-&-\\
MAR \cite{yu2019unsupervised} & CVPR 19&MSMT&67.7&81.9&-&40.0&MSMT& 67.1&79.8&-&48.0\\
ECN \cite{zhong2019invariance}&CVPR 19&Duke&75.1&87.6&91.6&43.0&Market&63.3&75.8&80.4&40.4\\
\hline
Ours&This paper&Duke&\bf 84.1&\bf 92.8&\bf 95.4&\bf 63.8&Market&\bf 74.0&\bf 83.7&\bf 87.4&\bf 54.4\\
\hline
\end{tabularx}
\end{center}
\end{table*}

\textbf{Benefit of exemplar memory.} In Table~\ref{tabel:memory and GPP}, we validate the effectiveness of the exemplar memory. When training our model without memory, we enforce the invariance learning within a mini-batch. Specifically, the inputs of the target mini-batch are composed of the target samples, corresponding CamStyle samples and corresponding $k$-nearest neighbors. Without the memory, we could not employ GPP in our model. Therefore, for a fair comparison, we train our memory based model without GPP, \textit{i.e.} select the reliable neighbors $\mathcal{K}(x^t_i)$ by vanilla neighbor selection (VNS) method. As shown in Table ~\ref{tabel:memory and GPP}, the exemplar memory based method clearly outperforms the mini-batch based method. This demonstrates the effectiveness of leveraging relations among whole datasets by memory. It is noteworthy that using the exemplar memory introduces limited additional training time ($\approx$ + 0.02 s/iter) and GPU memory ($\approx$ + 200 MB) compared to using the mini-batch.

\begin{figure}[!t]
\centering
\includegraphics[width=\linewidth]{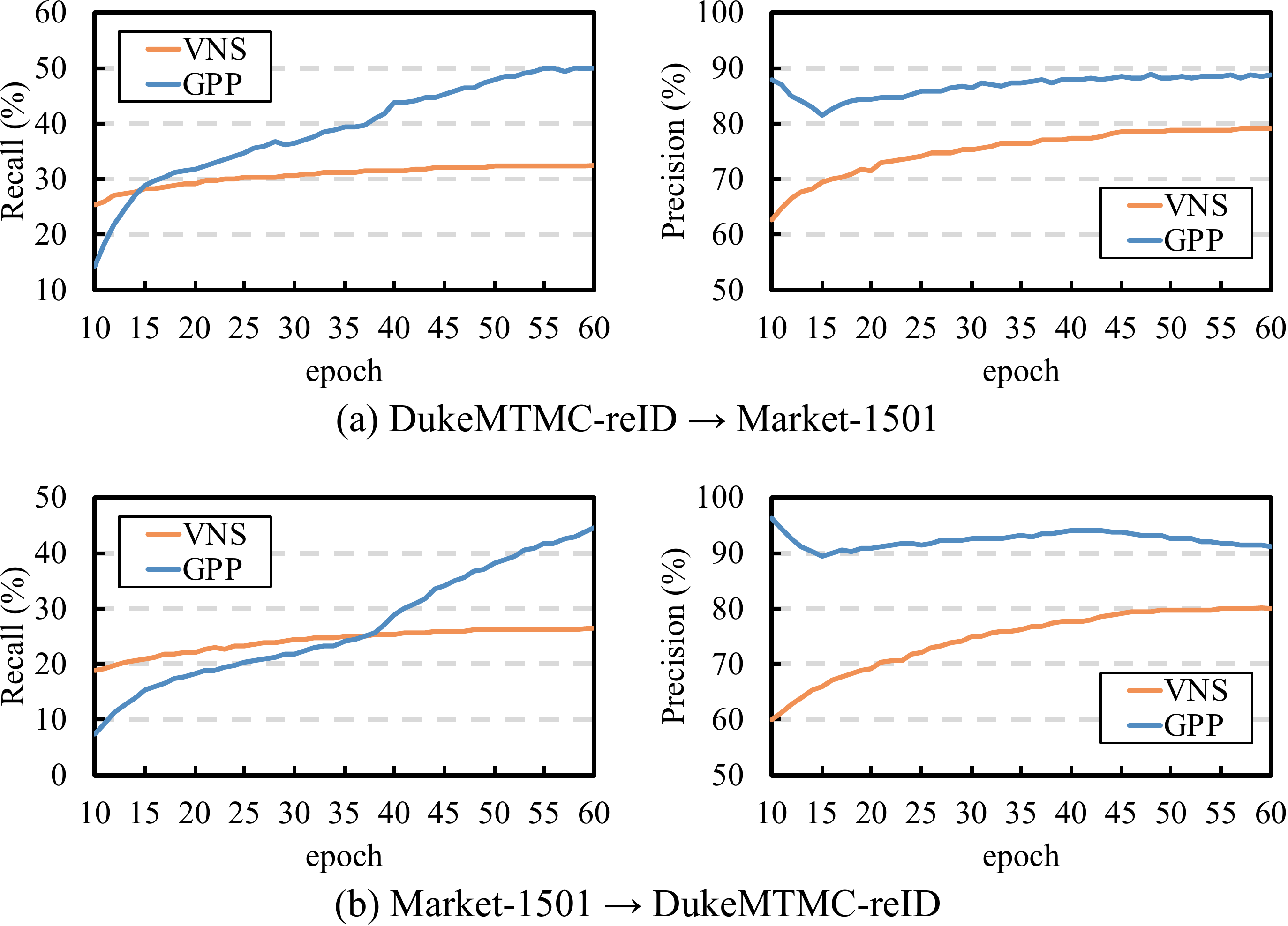}
\caption{The curve of selected reliable neighbors in (a) recall and (b) precision throughout the training. Results are compared between vanilla neighbor selection (VNS) and graph-based positive prediction (GPP).}
\label{fig:recall}
\end{figure}

\textbf{Effectiveness of graph-based positive prediction.} In Table~\ref{tabel:memory and GPP}, we report the results of training our method with and without graph-based positive prediction (GPP). Without GPP, our method reduces to training with vanilla neighbor selection (VNS), \textit{i.e.} directly select $k$-nearest neighbors from memory as reliable neighbors $\mathcal{K}(x^t_i)$. It is clearly that, adding GPP into the system significantly improves the accuracy, requiring extra 0.11 s/iter training time and 2,800 MB GPU memory.

In Table~\ref{tabel:neighor selection}, we also evaluate two other neighbor selection methods that are constructed as follows:

\begin{itemize}
    \item Variant VNS: We train a positive prediction classifier as the same way as GPP, but without using graph convolution layers to update features of candidate samples. $\mathcal{K}(x^t_i)$ is selected according to the predicted positive probability and threshold $\mu$. We set $\mu=0.9$ for variant VNS.
    
    \item Variant GPP: We use the same architecture as the proposed GPP. The top-$k$ samples are selected as $\mathcal{K}(x^t_i)$, according to the positive prediction of GPP. We set $k=8$ for variant GPP.
\end{itemize}

Table~\ref{tabel:neighor selection} shows that: \textbf{(1)} GPP based methods consistently outperforms VNS based methods, whether selecting $\mathcal{K}(x^t_i)$ by top-$k$ samples or threshold of predicted positive probability. This demonstrates the effectiveness of GPP, which updates features with the neighbor relations. \textbf{(2)} For GPP based methods, it is better to selecting reliable neighbors based on threshold than fix top-$k$ samples. Because the number of positive samples of each sample is very different, using fix top-$k$ samples might include negative samples that have low positive probability or ignore positive samples that have high positive probability. While threshold based strategy is more flexible and will always select samples with high positive probability. \textbf{(3)} The improvement of threshold based strategy is limited or even negative for the VNS method. Because the inputs of positive prediction classifier are unrelated features, which do not consider the relations among candidates. In this case, the positive prediction classifier fails to refine the similarities between the input and candidates.

To further validate the effectiveness of GPP, we evaluate the selected reliable neighbors $\mathcal{K}(x^t_i)$ throughout the training for VNS and GPP. Two metrics are applied: recall and precision. Recall is the fraction of true positive samples in $\mathcal{K}(x^t_i)$ over the total amount of positive samples in the dataset. Precision is the fraction of true positive samples in $\mathcal{K}(x^t_i)$ among the number of samples in $\mathcal{K}(x^t_i)$. Fig.~\ref{fig:recall} shows that: 1) GPP consistently produces higher precision than VNS method. 2) The recall of GPP is lower than VNS in the early training epochs. However, the recall of GPP consistently grows with the training epochs and is largely higher than that of VNS in later training epochs. This confirms the superiority of GPP over VNS in terms of reliable neighbor selection and adaptation accuracy.

\subsection{Comparison with State-of-the-art Methods}

In Table~\ref{tabel:sota-m-d} and Table~\ref{tabel:MSMT17}, we compare our approach with state-of-the-art unsupervised learning methods when tested on Market-1501, DukeMTMC-reID and MSMT17.

\textbf{SOTA on Market-1501 and DukeMTMC-reID.} 
Table~\ref{tabel:sota-m-d} reports the comparisons when tested on Market-1501 and DukeMTMC-reID. We compare with two hand-crafted feature based methods without transfer learning: LOMO \cite{liao2015lomo} and BOW \cite{zheng2015scalable}, four unsupervised methods that use a labeled source data to initialize the model but ignore the labeled source data during learning feature for the target domain: CAMEL \cite{yu2017cross}, DECAMEL \cite{yu2018unsupervised}, UMDL \cite{peng2016unsupervised} and PUL \cite{fan2017pul}, and nine unsupervised domain adaptation approaches: PTGAN \cite{wei2018person}, SPGAN \cite{deng2018image}, MMFA \cite{lin2018multibmvc}, TJ-AIDL \cite{wang2018reid}, DASy \cite{Bak_2018_ECCV}, MAR \cite{yu2019unsupervised}, CamStyle \cite{zhong2019camstyle}, HHL \cite{Zhong_2018_ECCV} and ECN \cite{zhong2019invariance}. 

We first compare with hand-crafted feature based methods, which neither require learning on labeled source set nor unlabeled target set. These two hand-crafted features have demonstrated the effectiveness on small datasets, but fail to produce competitive results on large-scale datasets. For example, the rank-1 accuracy of LOMO is 12.3\% when tested on DukeMTMC-reID, which is much lower than transferring learning based methods. 

Next, we compare with four unsupervised methods. Benefit from initializing model from the labeled source data and learning with unlabeled target data, the results of these three unsupervised approaches are commonly superior to hand-crafted methods. Such as, PUL obtains rank-1 accuracy of 45.5\% when using DukeMTMC-reID as source set and tested on Market-1501, surpassing BOW by 9.7\% in rank-1 accuracy.

\begin{table}[!t]
\caption{\label{tabel:MSMT17} Unsupervised/semi-supervised person re-ID performance comparison with state-of-the-art methods on MSMT17. \textbf{*}: Using within-camera identity annotations of target dataset. \textbf{Multi}: a combination of seven datasets.}
\vspace{-.2in}
\begin{center}
\newcolumntype{C}{>{\centering\arraybackslash}X}%
\newcolumntype{R}{>{\raggedleft\arraybackslash}X}%
\begin{tabularx}{\linewidth}{l|c|c|CCcC}
\hline
\multicolumn{1}{l|}{\multirow{2}{*}{Methods}}&\multirow{2}{*}{Reference}&\multirow{2}{*}{Source} &\multicolumn{4}{c}{MSMT17}\\
\cline{4-7}
&&&{R-1}&{R-5}&{R-10}&{mAP}\\
\hline
\hline
TAUDL \cite{li2018unsupervised}&ECCV 18&\multirow{2}{*}{MSMT*}& 28.4&-&-&12.5\\
UTAL \cite{li2019unsupervised}&TPAMI 19&&31.4&-&-&13.1\\
\hline
DECA \cite{yu2018unsupervised} &TPAMI 19&Multi&30.3&-&-&11.1 \\
\hline
PTGAN \cite{wei2018person}&CVPR 18&\multirow{4}{*}{Market}& 10.2&-& 24.4& 2.9\\
ECN \cite{zhong2019invariance}&CVPR 19&&25.3&36.3&42.1&8.5\\
Ours &This paper &&\bf 40.4&\bf 53.1&\bf 58.7&\bf 15.2\\
\hline
PTGAN \cite{wei2018person}&CVPR 18&\multirow{4}{*}{Duke}&11.8&-&27.4& 3.3\\
ECN \cite{zhong2019invariance}& CVPR 19&& 30.2&41.5&46.8&10.2\\
Ours &This paper&&\bf 42.5&\bf 55.9&\bf 61.5&\bf 16.0\\
\hline
\end{tabularx}
\end{center}
\end{table}

Finally, we compare with the domain adaptation based approaches, which produce state-of-the-art results. As can be seen, our approach outperforms them by a large margin on both datasets. Specifically, our method achieves \textbf{rank-1 accuracy = 84.1\%} and \textbf{mAP = 63.8\%} when using DukeMTMC-reID as the source set and tested on Market-1501, and, obtains \textbf{rank-1 accuracy = 74.0\%} and \textbf{mAP = 54.4\%} vice-versa.
It is worth noting that our approach significantly outperforms MAR \cite{yu2019unsupervised} which uses a larger dataset (MSMT17) as the source domain. Compared to the current best method (ECN \cite{zhong2019invariance}), our method surpasses ECN by 20\% and 14.4\% in mAP when tested on Market-1501 and DukeMTMC-reID, respectively.

\textbf{SOTA on MSMT17.} We also evaluate our approach on a larger and more challenging dataset, \emph{i.e.}, MSMT17. Since it is a newly released dataset, there are only three unsupervised domain adaptation methods (PTGAN \cite{wei2018person}, DECA \cite{yu2019unsupervised}, and ECN \cite{zhong2019invariance}) reported results on MSMT17. In addition, we further compare with two semi-supervised methods, TAUDL \cite{li2018unsupervised} and UTAL \cite{li2019unsupervised}, which use within-camera identity annotations of the target domain. As shown in Table~\ref{tabel:MSMT17}, our approach clearly exceeds both unsupervised domain adaptation methods and semi-supervised methods. For instance, our method produces \textbf{rank-1 accuracy = 42.5\%} and \textbf{mAP = 16.0\%} when using DukeMTMC-reID as the source set, which is higher than ECN by 12.3\% in rank-1 accuracy and by 5.8\% in mAP. Similar superiority of our method can be observed when using the Market-1501 as the source domain.

\section{Conclusion}

In this paper, we propose an exemplar memory based unsupervised domain adaptation (UDA) framework for person re-ID task. With the exemplar memory, we can directly evaluate the relationships between target samples. And thus we could effectively enforce the underlying invariance constraints of the target domain into the network training process. Moreover, the memory enables us to design a graph-based positive prediction (GPP) method that can infer reliable neighbors from candidate neighbors. Experiment demonstrates the effectiveness of the invariance learning, memory and GPP for improving the transferable ability of deep re-ID model. Our approach produces new state of the art in UDA accuracy on three large-scale domains.



{\small
\bibliographystyle{ieee}
\bibliography{egbib}

\begin{thebibliography}{10}\itemsep=-1pt

\bibitem{Bak_2018_ECCV}
Slawomir Bak, Peter Carr, and Jean-Francois Lalonde.
\newblock Domain adaptation through synthesis for unsupervised person
  re-identification.
\newblock In {\em Proc. ECCV}, 2018.

\bibitem{bousmalis2016domain}
Konstantinos Bousmalis, George Trigeorgis, Nathan Silberman, Dilip Krishnan,
  and Dumitru Erhan.
\newblock Domain separation networks.
\newblock In {\em Proc. NIPS}, 2016.

\bibitem{bruna2014spectral}
Joan Bruna, Wojciech Zaremba, Arthur Szlam, and Yann LeCun.
\newblock Spectral networks and locally connected networks on graphs.
\newblock In {\em Proc. ICLR}, 2014.

\bibitem{busto2017open-set}
Pau~Panareda Busto and Juergen Gall.
\newblock Open set domain adaptation.
\newblock In {\em Proc. ICCV}, 2017.

\bibitem{chen2018deep}
Yanbei Chen, Xiatian Zhu, and Shaogang Gong.
\newblock Deep association learning for unsupervised video person
  re-identification.
\newblock In {\em Proc. BMVC}, 2018.

\bibitem{stargan}
Yunjey Choi, Minje Choi, Munyoung Kim, Jung-Woo Ha2~Sunghun Kim, and Jaegul
  Choo.
\newblock Stargan: Unified generative adversarial networks for multi-domain
  image-to-image translation.
\newblock In {\em Proc. CVPR}, 2018.

\bibitem{defferrard2016convolutional}
Micha{\"e}l Defferrard, Xavier Bresson, and Pierre Vandergheynst.
\newblock Convolutional neural networks on graphs with fast localized spectral
  filtering.
\newblock In {\em Advances in neural information processing systems}, pages
  3844--3852, 2016.

\bibitem{deng2009imagenet}
Jia Deng, Wei Dong, Richard Socher, Li-Jia Li, Kai Li, and Li Fei-Fei.
\newblock Imagenet: A large-scale hierarchical image database.
\newblock In {\em Proc. CVPR}, 2009.

\bibitem{deng2018image}
Weijian Deng, Liang Zheng, Qixiang Ye, Guoliang Kang, Yi Yang, and Jianbin
  Jiao.
\newblock Image-image domain adaptation with preserved self-similarity and
  domain-dissimilarity for person re-identification.
\newblock In {\em Proc. CVPR}, 2018.

\bibitem{fan2017pul}
Hehe Fan, Liang Zheng, Chenggang Yan, and Yi Yang.
\newblock Unsupervised person re-identification: Clustering and fine-tuning.
\newblock {\em ACM TOMM}, 2018.

\bibitem{graves2014neural}
Alex Graves, Greg Wayne, and Ivo Danihelka.
\newblock Neural turing machines.
\newblock {\em arXiv preprint arXiv:1410.5401}, 2014.

\bibitem{gray2008viewpoint}
Douglas Gray and Hai Tao.
\newblock Viewpoint invariant pedestrian recognition with an ensemble of
  localized features.
\newblock In {\em Proc. ECCV}, 2008.

\bibitem{gretton2007kernel}
Arthur Gretton, Karsten~M Borgwardt, Malte Rasch, Bernhard Sch{\"o}lkopf, and
  Alex~J Smola.
\newblock A kernel method for the two-sample-problem.
\newblock In {\em Proc. NIPS}, 2007.

\bibitem{hadsell2006contrastive}
Raia Hadsell, Sumit Chopra, and Yann LeCun.
\newblock Dimensionality reduction by learning an invariant mapping.
\newblock In {\em Proc. CVPR}, 2006.

\bibitem{hamilton2017inductive}
Will Hamilton, Zhitao Ying, and Jure Leskovec.
\newblock Inductive representation learning on large graphs.
\newblock In {\em Advances in Neural Information Processing Systems}, pages
  1024--1034, 2017.

\bibitem{he2015PReLU}
Kaiming He, Xiangyu Zhang, Shaoqing Ren, and Jian Sun.
\newblock Delving deep into rectifiers: Surpassing human-level performance on
  imagenet classification.
\newblock In {\em Proc. ICCV}, pages 1026--1034, 2015.

\bibitem{resnet}
Kaiming He, Xiangyu Zhang, Shaoqing Ren, and Jian Sun.
\newblock Deep residual learning for image recognition.
\newblock In {\em Proc. CVPR}, 2016.

\bibitem{henaff2015deep}
Mikael Henaff, Joan Bruna, and Yann LeCun.
\newblock Deep convolutional networks on graph-structured data.
\newblock {\em arXiv preprint arXiv:1506.05163}, 2015.

\bibitem{hermans2017defense}
Alexander Hermans, Lucas Beyer, and Bastian Leibe.
\newblock In defense of the triplet loss for person re-identification.
\newblock {\em arXiv preprint arXiv:1703.07737}, 2017.

\bibitem{ioffe2015batch}
Sergey Ioffe and Christian Szegedy.
\newblock Batch normalization: accelerating deep network training by reducing
  internal covariate shift.
\newblock In {\em Proc. ICML}, 2015.

\bibitem{kipf2016semi}
Thomas~N Kipf and Max Welling.
\newblock Semi-supervised classification with graph convolutional networks.
\newblock In {\em Proc. ICLR}, 2017.

\bibitem{li2018unsupervised}
Minxian Li, Xiatian Zhu, and Shaogang Gong.
\newblock Unsupervised person re-identification by deep learning tracklet
  association.
\newblock In {\em Proceedings of the European Conference on Computer Vision
  (ECCV)}, pages 737--753, 2018.

\bibitem{li2019unsupervised}
Minxian Li, Xiatian Zhu, and Shaogang Gong.
\newblock Unsupervised tracklet person re-identification.
\newblock {\em IEEE transactions on pattern analysis and machine intelligence},
  2019.

\bibitem{Li_2018_CVPR}
Wei Li, Xiatian Zhu, and Shaogang Gong.
\newblock Harmonious attention network for person re-identification.
\newblock In {\em Proc. CVPR}, 2018.

\bibitem{liao2015lomo}
Shengcai Liao, Yang Hu, Xiangyu Zhu, and Stan~Z Li.
\newblock Person re-identification by local maximal occurrence representation
  and metric learning.
\newblock In {\em Proc. CVPR}, 2015.

\bibitem{lin2013network}
Min Lin, Qiang Chen, and Shuicheng Yan.
\newblock Network in network.
\newblock In {\em Proc. ICLR}, 2014.

\bibitem{lin2018multibmvc}
Shan Lin, Haoliang Li, Chang-Tsun Li, and Alex~Chichung Kot.
\newblock Multi-task mid-level feature alignment network for unsupervised
  cross-dataset person re-identification.
\newblock In {\em Prco. BMVC}, 2018.

\bibitem{long2015learning}
Mingsheng Long, Yue Cao, Jianmin Wang, and Michael~I Jordan.
\newblock Learning transferable features with deep adaptation networks.
\newblock In {\em Proc. ICML}, 2015.

\bibitem{Yawei2019Taking}
Yawei Luo, Liang Zheng, Tao Guan, Junqing Yu, and Yi Yang.
\newblock Taking a closer look at domain shift: Category-level adversaries for
  semantics consistent domain adaptation.
\newblock In {\em Proc. CVPR}, 2019.

\bibitem{monti2017geometric}
Federico Monti, Davide Boscaini, Jonathan Masci, Emanuele Rodola, Jan Svoboda,
  and Michael~M Bronstein.
\newblock Geometric deep learning on graphs and manifolds using mixture model
  cnns.
\newblock In {\em Proceedings of the IEEE Conference on Computer Vision and
  Pattern Recognition}, pages 5115--5124, 2017.

\bibitem{nair2010relu}
Vinod Nair and Geoffrey~E Hinton.
\newblock Rectified linear units improve restricted boltzmann machines.
\newblock In {\em Proc. ICML}, 2010.

\bibitem{niepert2016learning}
Mathias Niepert, Mohamed Ahmed, and Konstantin Kutzkov.
\newblock Learning convolutional neural networks for graphs.
\newblock In {\em International conference on machine learning}, pages
  2014--2023, 2016.

\bibitem{peng2016unsupervised}
Peixi Peng, Tao Xiang, Yaowei Wang, Massimiliano Pontil, Shaogang Gong, Tiejun
  Huang, and Yonghong Tian.
\newblock Unsupervised cross-dataset transfer learning for person
  re-identification.
\newblock In {\em Proc. CVPR}, 2016.

\bibitem{ristani2016performance}
Ergys Ristani, Francesco Solera, Roger Zou, Rita Cucchiara, and Carlo Tomasi.
\newblock Performance measures and a data set for multi-target, multi-camera
  tracking.
\newblock In {\em Proc. ECCVW}, 2016.

\bibitem{saito2018open}
Kuniaki Saito, Shohei Yamamoto, Yoshitaka Ushiku, and Tatsuya Harada.
\newblock Open set domain adaptation by backpropagation.
\newblock In {\em Proc. ECCV}, 2018.

\bibitem{santoro2016meta}
Adam Santoro, Sergey Bartunov, Matthew Botvinick, Daan Wierstra, and Timothy
  Lillicrap.
\newblock Meta-learning with memory-augmented neural networks.
\newblock In {\em Proc. ICML}, 2016.

\bibitem{shen2018graph}
Yantao Shen, Hongsheng Li, Shuai Yi, Dapeng Chen, and Xiaogang Wang.
\newblock Person re-identification with deep similarity-guided graph neural
  network.
\newblock In {\em Proc. ECCV}, 2018.

\bibitem{sohn2019unsupervised}
Kihyuk Sohn, Wenling Shang, Xiang Yu, and Manmohan Chandraker.
\newblock Unsupervised domain adaptation for distance metric learning.
\newblock In {\em Proc. ICLR}, 2019.

\bibitem{srivastava2014dropout}
Nitish Srivastava, Geoffrey Hinton, Alex Krizhevsky, Ilya Sutskever, and Ruslan
  Salakhutdinov.
\newblock Dropout: a simple way to prevent neural networks from overfitting.
\newblock {\em JMLR}, 2014.

\bibitem{sukhbaatar2015end}
Sainbayar Sukhbaatar, Jason Weston, Rob Fergus, et~al.
\newblock End-to-end memory networks.
\newblock In {\em Proc. NIPS}, 2015.

\bibitem{sun2019dissecting}
Xiaoxiao Sun and Liang Zheng.
\newblock Dissecting person re-identification from the viewpoint of viewpoint.
\newblock In {\em Proc. CVPR}, 2019.

\bibitem{sun2018beyond}
Yifan Sun, Liang Zheng, Yi Yang, Qi Tian, and Shengjin Wang.
\newblock Beyond part models: Person retrieval with refined part pooling (and a
  strong convolutional baseline).
\newblock In {\em Proc. ECCV}, 2018.

\bibitem{tzeng2017adversarial}
Eric Tzeng, Judy Hoffman, Kate Saenko, and Trevor Darrell.
\newblock Adversarial discriminative domain adaptation.
\newblock In {\em Proc. CVPR}, 2017.

\bibitem{vinyals2016matching}
Oriol Vinyals, Charles Blundell, Tim Lillicrap, Daan Wierstra, et~al.
\newblock Matching networks for one shot learning.
\newblock In {\em Proc. NIPS}, 2016.

\bibitem{wang2018reid}
Jingya Wang, Xiatian Zhu, Shaogang Gong, and Wei Li.
\newblock Transferable joint attribute-identity deep learning for unsupervised
  person re-identification.
\newblock In {\em Proc. CVPR}, 2018.

\bibitem{wang2019linkage}
Zhongdao Wang, Liang Zheng, Yali Li, and Shengjin Wang.
\newblock Linkage based face clustering via graph convolution network.
\newblock In {\em Proc. CVPR}, 2019.

\bibitem{wei2018person}
Longhui Wei, Shiliang Zhang, Wen Gao, and Qi Tian.
\newblock Person transfer gan to bridge domain gap for person
  re-identification.
\newblock In {\em Proc. CVPR}, 2018.

\bibitem{MemoryNetworks2015}
Jason Weston, Sumit Chopra, and Antoine Bordes.
\newblock Memory networks.
\newblock In {\em Proc. ICLR}, 2015.

\bibitem{wu2018long}
Chao-Yuan Wu, Christoph Feichtenhofer, Haoqi Fan, Kaiming He, Philipp
  Kr{\"a}henb{\"u}hl, and Ross Girshick.
\newblock Long-term feature banks for detailed video understanding.
\newblock In {\em Proc. CVPR}, 2019.

\bibitem{wu2018improving}
Zhirong Wu, Alexei~A Efros, and Stella~X Yu.
\newblock Improving generalization via scalable neighborhood component
  analysis.
\newblock In {\em Proc. ECCV}, 2018.

\bibitem{wu2018unsupervised}
Zhirong Wu, Yuanjun Xiong, X~Yu Stella, and Dahua Lin.
\newblock Unsupervised feature learning via non-parametric instance
  discrimination.
\newblock In {\em Proc. CVPR}, 2018.

\bibitem{xiao2017joint}
Tong Xiao, Shuang Li, Bochao Wang, Liang Lin, and Xiaogang Wang.
\newblock Joint detection and identification feature learning for person
  search.
\newblock In {\em Proc. CVPR}, 2017.

\bibitem{yan2017mind}
Hongliang Yan, Yukang Ding, Peihua Li, Qilong Wang, Yong Xu, and Wangmeng Zuo.
\newblock Mind the class weight bias: Weighted maximum mean discrepancy for
  unsupervised domain adaptation.
\newblock In {\em Proc. CVPR}, 2017.

\bibitem{yang2018leveraging}
Fengxiang Yang, Zhun Zhong, Zhiming Luo, Sheng Lian, and Shaozi Li.
\newblock Leveraging virtual and real person for unsupervised person
  re-identification.
\newblock {\em arXiv preprint arXiv:1811.02074}, 2018.

\bibitem{yu2017cross}
Hong-Xing Yu, Ancong Wu, and Wei-Shi Zheng.
\newblock Cross-view asymmetric metric learning for unsupervised person
  re-identification.
\newblock In {\em Proc. ICCV}, 2017.

\bibitem{yu2018unsupervised}
Hong-Xing Yu, Ancong Wu, and Wei-Shi Zheng.
\newblock Unsupervised person re-identification by deep asymmetric metric
  embedding.
\newblock {\em IEEE TPAMI}, 2019.

\bibitem{yu2019unsupervised}
Hong-Xing Yu, Wei-Shi Zheng, Ancong Wu, Xiaowei Guo, Shaogang Gong, and
  Jian-Huang Lai.
\newblock Unsupervised person re-identification by soft multilabel learning.
\newblock In {\em Proc. CVPR}, 2019.

\bibitem{zheng2015scalable}
Liang Zheng, Liyue Shen, Lu Tian, Shengjin Wang, Jingdong Wang, and Qi Tian.
\newblock Scalable person re-identification: A benchmark.
\newblock In {\em Proc. ICCV}, 2015.

\bibitem{zheng2016personsurvery}
Liang Zheng, Yi Yang, and Alexander~G Hauptmann.
\newblock Person re-identification: Past, present and future.
\newblock {\em arXiv}, 2016.

\bibitem{zheng2019joint}
Zhedong Zheng, Xiaodong Yang, Zhiding Yu, Liang Zheng, Yi Yang, and Jan Kautz.
\newblock Joint discriminative and generative learning for person
  re-identification.
\newblock In {\em Proc. CVPR}, 2019.

\bibitem{zheng2017unlabeled}
Zhedong Zheng, Liang Zheng, and Yi Yang.
\newblock Unlabeled samples generated by gan improve the person
  re-identification baseline in vitro.
\newblock In {\em Proc. ICCV}, 2017.

\bibitem{zhong2017re}
Zhun Zhong, Liang Zheng, Donglin Cao, and Shaozi Li.
\newblock Re-ranking person re-identification with k-reciprocal encoding.
\newblock In {\em Proc. CVPR}, 2017.

\bibitem{zhong2017random}
Zhun Zhong, Liang Zheng, Guoliang Kang, Shaozi Li, and Yi Yang.
\newblock Random erasing data augmentation.
\newblock {\em arXiv}, 2017.

\bibitem{Zhong_2018_ECCV}
Zhun Zhong, Liang Zheng, Shaozi Li, and Yi Yang.
\newblock Generalizing a person retrieval model hetero- and homogeneously.
\newblock In {\em Proc. ECCV}, 2018.

\bibitem{zhong2019invariance}
Zhun Zhong, Liang Zheng, Zhiming Luo, Shaozi Li, and Yi Yang.
\newblock Invariance matters: Exemplar memory for domain adaptive person
  re-identiﬁcation.
\newblock In {\em The IEEE Conference on Computer Vision and Pattern
  Recognition (CVPR)}. IEEE, 2019.

\bibitem{zhong2018camera}
Zhun Zhong, Liang Zheng, Zhedong Zheng, Shaozi Li, and Yi Yang.
\newblock Camera style adaptation for person re-identification.
\newblock In {\em Proc. CVPR}, 2018.

\bibitem{zhong2019camstyle}
Zhun Zhong, Liang Zheng, Zhedong Zheng, Shaozi Li, and Yi Yang.
\newblock Camstyle: A novel data augmentation method for person
  re-identification.
\newblock {\em IEEE TIP}, 2019.

\bibitem{zhu2017cyclegan}
Jun-Yan Zhu, Taesung Park, Phillip Isola, and Alexei~A Efros.
\newblock Unpaired image-to-image translation using cycle-consistent
  adversarial networks.
\newblock In {\em Proc. ICCV}, 2017.

\end{thebibliography}
}

%










\end{document}